\documentclass[journal]{IEEEtran}

\usepackage{amsmath,amsfonts,amsthm}
\usepackage{verbatim}
\usepackage{graphicx} 
\usepackage{enumitem} 
\usepackage{cite}
\usepackage{float}
\usepackage{siunitx}
\usepackage[T1]{fontenc}
\usepackage{tikz}
\usepackage[flushleft]{threeparttable}
\usepackage[normalem]{ulem}
\usepackage{booktabs}
\usepackage{multirow}
\usepackage{makecell}
\usepackage{xcolor}
\usepackage{authblk} 
\usepackage[pdfpagemode=UseNone]{hyperref} 
\hypersetup{
    pdftoolbar=true,        
    pdfmenubar=true,        
    pdffitwindow=false,     
    pdfstartview={FitH},    
    pdftitle={LiDAR Localizability Aware Constrained Optimization for Robust Robot Pose Estimation},    
    pdfauthor={Turcan Tuna},     
    pdfsubject={IEEE TRO Journal},   
    pdfcreator={Turcan Tuna},   
    pdfproducer={Turcan Tuna}, 
    colorlinks=true,
    linkcolor=blue,
    linkbordercolor=white,
    filecolor=magenta,
    citecolor=red,
    urlcolor=cyan
    }
\usetikzlibrary{calc}
\usepackage[export]{adjustbox}

\IEEEoverridecommandlockouts

\def\BibTeX{{\rm B\kern-.05em{\sc i\kern-.025em b}\kern-.08em
    T\kern-.1667em\lower.7ex\hbox{E}\kern-.125emX}}
\usepackage{balance}

\usepackage{subfiles}
\usepackage{ifthen}
\newboolean{printBibInSubfiles}
\setboolean{printBibInSubfiles}{true} 
\def\bib{\ifthenelse{\boolean{printBibInSubfiles}}
        {\bibliographystyle{IEEEtran}\bibliography{references.bib}}
} 

\setboolean{printBibInSubfiles}{false}

\definecolor{darkgreen}{rgb}{0.0, 0.5, 0.0}
\definecolor{alizarin}{rgb}{0.82, 0.1, 0.26}
\definecolor{bittersweet}{rgb}{0.94, 0.5, 0.5} 
\definecolor{bananamania}{rgb}{0.93, 0.86, 0.51}
\definecolor{applegreen}{rgb}{0.67, 0.88, 0.69}
\definecolor{bluebell}{rgb}{0.64, 0.64, 0.82}
\definecolor{bisque}{rgb}{0.95, 0.9, 0.67}

\newcommand{\purpleline}{\raisebox{-1pt}{\tikz{\draw[-,bluebell,solid,line width = 6.8pt](0,0) -- (4mm,0mm);}}}
\newcommand{\yellowline}{\raisebox{-1pt}{\tikz{\draw[-,bananamania,solid,line width = 6.8pt](0,0) -- (4mm,0mm);}}}
\newcommand{\redcircle}{\raisebox{-1pt}{\tikz\draw[red,fill=red] (0,0) circle (0.75ex);}}
\newcommand{\greenline}{\raisebox{-1pt}{\tikz{\draw[-,applegreen,solid,line width = 6.8pt](0,0) -- (4mm,0mm);}}}
\newcommand{\redline}{\raisebox{-1pt}{\tikz{\draw[-,bittersweet,solid,line width = 6.8pt](0,0) -- (4mm,0mm);}}}
\newcommand{\greenlinet}{\raisebox{-1pt}{\tikz{\draw[-,bisque,solid,line width = 6.8pt](0,0) -- (4mm,0mm);}}}

\title{X-ICP: Localizability-Aware LiDAR Registration for Robust Localization in Extreme Environments} 

\author{
Turcan Tuna$^{\dagger,\ddagger}$,
Julian Nubert$^{\dagger}$,
Yoshua Nava$^{\ddagger}$,
Shehryar Khattak$^{\dagger}$,
Marco Hutter$^{\dagger}$
\thanks{This work is supported in part by the EU Horizon 2020 programme grant agreement No. 852044, 101016970, and 101070405, EU Horizon 2021 programme grant agreement No. 101070596, the NCCR digital fabrication and robotics, the ETH Zurich Research Grant No. 21-1 ETH-27, the SNF project No. 188596, and the Max Planck ETH Center for Learning Systems.}
\thanks{$^{\dagger}$The authors are with the Robotics Systems Lab, ETH Z\"urich.}
\thanks{$^{\ddagger}$The authors are with the ANYbotics A.G.}
}

\usepackage{fancyhdr}
\newcommand{\mytitle}{\textbf{Accepted version.} To appear in \textit{IEEE Transactions on Robotics, Vol. 40, 2024.}.  DOI:
10.1109/TRO.2023.3335691\\
\copyright 2024 IEEE. Personal use of this material is permitted.
Permission from IEEE must be obtained for all other uses, in any current or future media, including reprinting/republishing this material for advertising or promotional purposes, creating new collective works, for resale or redistribution to servers or lists, or reuse of any copyrighted component of this work in other works.} 
\begin{document}

\maketitle

\thispagestyle{fancy}

\begin{abstract}
Modern robotic systems are required to operate in challenging environments, which demand reliable localization under challenging conditions. LiDAR-based localization methods, such as the Iterative Closest Point (ICP) algorithm, can suffer in geometrically uninformative environments that are known to deteriorate point cloud registration performance and push optimization toward divergence along weakly constrained directions. To overcome this issue, this work proposes i) a robust fine-grained localizability detection module, and ii) a localizability-aware constrained ICP optimization module, which couples with the localizability detection module in a unified manner. The proposed localizability detection is achieved by utilizing the correspondences between the scan and the map to analyze the alignment strength against the principal directions of the optimization as part of its fine-grained LiDAR localizability analysis.  In the second part, this localizability analysis is then integrated into the scan-to-map point cloud registration to generate drift-free pose updates by enforcing controlled updates or leaving the degenerate directions of the optimization unchanged. The proposed method is thoroughly evaluated and compared to state-of-the-art methods in simulated and real-world experiments\footnote{\url{https://youtu.be/SviLl7q69aA}\label{foot:supplementary_material}}, demonstrating the performance and reliability improvement in LiDAR-challenging environments. In all experiments, the proposed framework demonstrates accurate and generalizable localizability detection and robust pose estimation without environment-specific parameter tuning.
\end{abstract}


\begin{figure}[!t]
\includegraphics[ width=\linewidth]{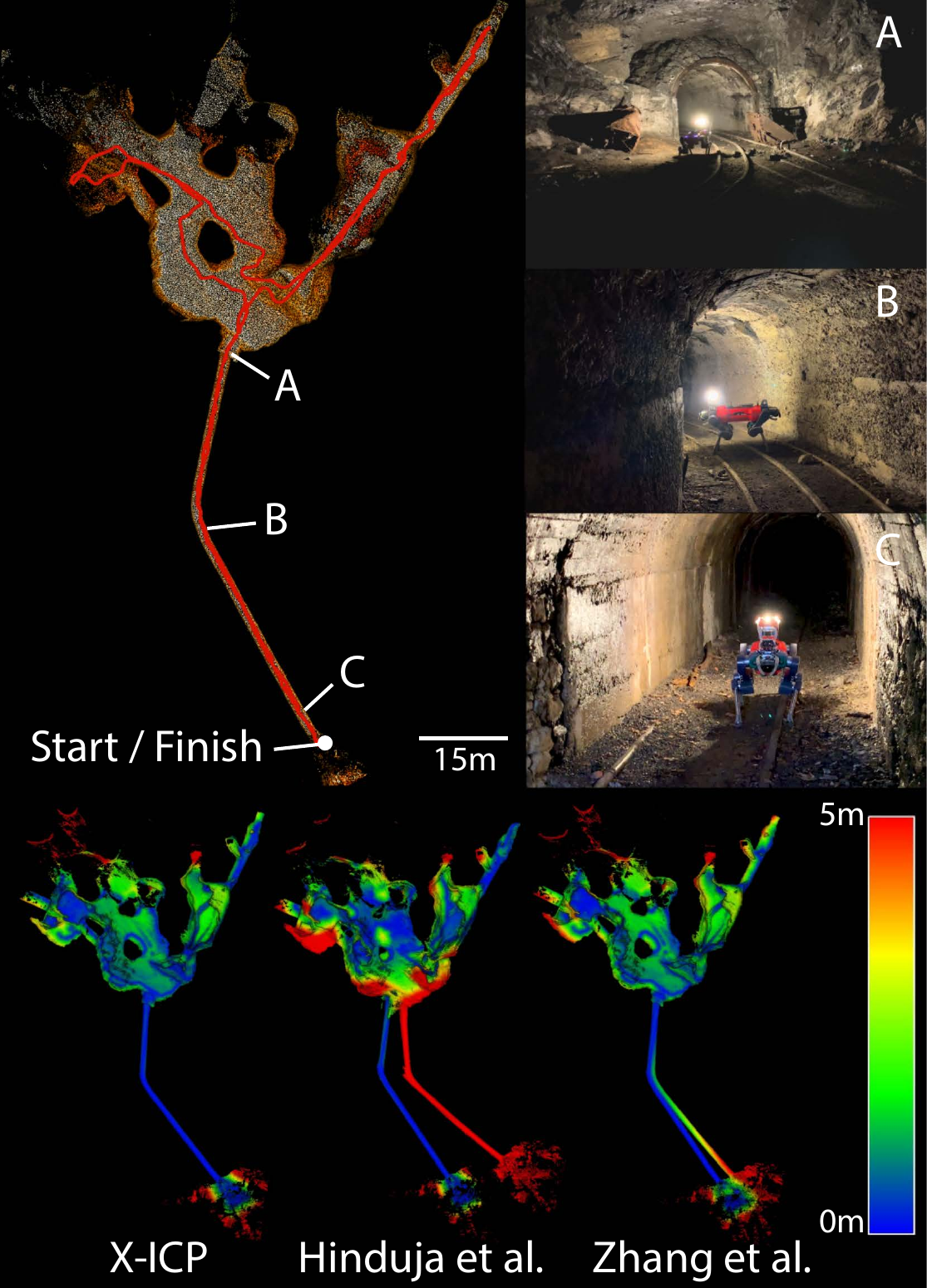}
\caption{\textbf{Top Row:} Ground truth map and path of the robot during the Seem\"uhle experiment. Certain sections of the environment are illustrated through real images. \textbf{Bottom Row:} Point cloud maps created using the proposed approach and compared against two state-of-the-art methods. The color bar indicates the point-to-point distance error with respect to the ground truth map.}
\label{fig:seemuhle_vlp16_result}
\vspace{-1.6em}
\end{figure}

\section{Introduction}\label{section:introduction}

\IEEEPARstart{R}ELIABLE robot pose estimation and map creation are core capabilities that enable mobile robots to operate autonomously. While it is possible to achieve robot localization with global positioning or fixed physical references (e.g., markers), these methods are expensive or difficult to scale to new environments. With the advent of increasingly capable exteroceptive sensors, the research community has focused on solving the \textit{simultaneous localization and mapping (SLAM)} problem with a variety of algorithms~\cite{slamOverview1, cadena2016past}. SLAM can be achieved with different types of sensors~\cite{slamOverview2}; however, this work focuses on LiDAR-based sensing due to its ability to provide reliable and accurate range measurements in the form of point clouds. The success of LiDAR-based approaches in competitions such as the KITTI~\cite{geiger2012we} and the recent HILTI~\cite{helmberger2022hilti} benchmarks support the strength of LiDAR-based approaches.

\paragraph{Point Cloud Registration} Most popular LiDAR-based SLAM algorithms perform point cloud registration using iterative error minimization techniques to estimate the correct pose difference between two point clouds, often referred to as \mbox{\textit{scan-to-scan}} registration. Similarly, point cloud registration can be done against a map, referred to as \mbox{\textit{scan-to-map}} registration, significantly reducing drift compared to \mbox{{scan-to-scan}}. In a typical LiDAR-based SLAM framework, \mbox{scan-to-map} registration is the core step where the pose of the incoming point cloud data in the map is calculated. To date, the most well-known point cloud registration method is the \mbox{\textit{iterative closest point (ICP)}} algorithm~\cite{pointToPointICP, point_to_plane}, still utilized by recent research~\cite{kiss_icp,jelavic2022open3d}.

\paragraph{Current Limitations} Although the ICP algorithm and its variants are commonly used, for practical applications, their performance is limited by four sources of error~\cite{censi, brossard}. These errors include \textit{a)} the risk of converging to a local minima, \textit{b)} the sensitivity to sensor measurement noise and bias, \textit{c)} utilization of an inaccurate transformation prior, and \textit{d)} the lack of geometric constraints provided by the environment for the underlying optimization problem. 
While the robotic community has developed solutions and frameworks that attenuate different error sources, the inaccurate transformation prior and the lack of geometric constraints can still cause modern LiDAR-based SLAM systems to fail when deployed in challenging environments. Although both problems are crucial, this work focuses on minimizing the adverse effect of the lack of geometrical constraints provided by the operational environment.
In self-symmetrical environments, environments with \textit{perceptual aliasing}, the geometric constraints along the axis of symmetry can be indistinguishable from noise, and the total number of unique constraints might be insufficient for optimization to converge. As a result, the optimization might converge to a noise-induced optimum~\cite{solRemap}, referred to as a \textit{ill-conditioned} or \textit{degenerate} solution. Tunnels, wide open spaces, tight corridors, and narrow doorways are instances of such degraded environments~\cite{selfSupervisedOdom}. An example of a real-world degenerate underground tunnel is shown in the top row of Fig.~\ref{fig:seemuhle_vlp16_result}. As discussed in recent research~\cite{nubert2022learning, zhen2017robust, tunnelLocalizability}, to enable robots to operate in these challenging scenarios, and to handle the solution ill-conditioning caused by degenerate planar or tunnel-like environments, localizability-awareness is required.

\paragraph{Proposed Approach} As a response to the LiDAR degeneracy problem, this work proposes a robust localizability-aware point cloud registration (ICP) framework that enables LiDAR-based SLAM systems to operate in featureless \textit{e\textbf{X}treme} environments, called \mbox{\textbf{X}-ICP}. The proposed framework, shown in Fig.~\ref{fig:highLevelOverview}, solves both the detection of LiDAR degeneracy and the mitigation of the adverse effects of this degeneracy on the optimization. The two sub-modules of the proposed approach are \textit{localizability detection} module, abbreviated as \textit{Loc.-Module}, and \textit{optimization} module, abbreviated as \textit{Opt.-Module}. The {Loc.-Module} utilizes the point and surface-normal correspondences between the scan and the map to analyze the contribution strength along the optimization's principal directions. In contrast to prior work~\cite{nubert2022learning}, the localizability detection is applied to the scan-to-map registration, and the localizability detection is done in the optimization \emph{eigenspace}, which allows the detection to be independent of the sensor orientation in the map. The resulting proposed solution enables reliable localizability detection in various environmental configurations, such as underground environments and large open spaces, without parameter tuning. Furthermore, the localizability detection is fine-grained and can classify the current optimization directions into three $\{$\textit{localizable}, \textit{partially-localizable}, and \textit{non-localizable}$\}$ categories.

Given the localizability information, the {Opt.-Module} calculates optimization constraints by salvaging the registration correspondences and employs these constraints in the low-level optimization for the point cloud registration. More specifically, this module leaves the point cloud registration initial guess unchanged in the \textit{non-localizable} directions and enforces controlled pose updates for \textit{partially-localizable} directions while having no effect on \textit{localizable} directions.
Notably, the Opt.-Module is independent of the ICP cost function and can be used independently in combination with other optimization-based systems since the adverse effects of the degenerate directions are mitigated by employing the localizability information within the iterative optimization problem. The result of this optimization utilizes the information contained in the correspondences while using the robot odometry information along the ill-conditioned directions. 

The proposed framework is tested extensively with multiple simulated environments and real-world missions in various environments and with different sensor setups. The field experiments and analyses suggest that the proposed framework can reliably detect localizability in diverse, challenging environments without requiring heuristic parameter tuning and achieves increased robustness and accuracy in degraded environments. The proposed framework consistently outperforms the state-of-the-art robotic approaches~\cite{solRemap,hinduja2019degeneracy} throughout all experiments performed in challenging and partly degenerate scenarios, as shown in Fig.~\ref{fig:seemuhle_vlp16_result}.

\paragraph{Contributions} The main contributions of this work are as follows:
\begin{itemize}
    \item A heuristic-free and multi-category localizability detection algorithm is developed to reliably identify the localizability state of the principal directions of the underlying point cloud registration optimization problem.
    \item A method to salvage scarce information from point correspondences is developed to utilize the available information in \textit{partially-localizable} optimization directions.
    \item A novel localizability-aware constrained ICP optimization module is developed for robust point cloud registration in degenerate scenarios.
    \item  A variety of experiments are conducted to evaluate the efficacy of the proposed framework and to compare the results against the state-of-the-art methods.
\end{itemize}
Additional content, data, and supplementary material are provided on the project page\footnote{Project page: \url{https://sites.google.com/leggedrobotics.com/x-icp}\label{foot:website}}.


\section{Related Work}\label{section:related_works}
LiDAR-based SLAM systems rely on point cloud registration methods to estimate the robot pose. A brief review of these methods is presented in Section~\ref{subsection:point_cloud_registration}. Furthermore, related work on degeneracy detection is discussed in Section~\ref{subsection:degeneracy_detection}, followed by a discussion of the approaches that constrain the underlying ill-conditioned optimization problem for point cloud registration in Section~\ref{subsection:constrainedOptInMapping}.
 
\subsection{Point Cloud Registration Methods}\label{subsection:point_cloud_registration}
Point cloud registration is considered a mature research field, and multiple unique approaches~\cite{NDT, generalizedICP, loam, behley2018efficient} are proposed to achieve robust, fast, and accurate pose estimation. Among these, the most widely used registration algorithm for LiDAR-based registration is the ICP algorithm~\cite{pomerleau2013applied}. The ICP algorithm iteratively finds the transformation between two point clouds given an initial transformation. This is achieved by minimizing a pre-defined cost function that measures the error between point pairs in source and target point clouds. Various cost functions have been proposed, such as the point-to-point~\cite{pointToPointICP}, point-to-plane~\cite{point_to_plane}, point-to-line~\cite{pointToLine}, point-to-Gaussian~\cite{point_to_gaussian}, and symmetric point-to-plane~\cite{symmetricPointToPlane}. Beyond that, works also combine different cost functions, statistical measures~\cite{generalizedICP, litamin2}, and even employ data-driven methods~\cite{selfSupervisedOdom} to achieve more robust registration. Despite these promising alternatives, the point-to-plane cost function~\cite{point_to_plane} is still among the preferred solutions for state-of-the-art robust LiDAR-based SLAM systems~\cite{wildCat,compSLAM,dareSLAM} due to its simplicity and effectiveness, as demonstrated in real-world deployments~\cite{slamOverview2,helmberger2022hilti}. 

In recent years, certifiable algorithms for robust point cloud registration have gained interest~\cite{teaser, wahba, lagrangianDuality}. These algorithms target to identify whether the optimization solution is globally optimal. Similarly, to perform reliable point cloud registration in the presence of outliers and strong noise, approaches such as~\cite{chebrolu, graduated} proposed using robust norms.
Nevertheless, conceptually, the problem definition of these works differs from the one of this work. In \mbox{X-ICP}, the goal is to identify possible degenerate cases \emph{before} the registration, while certifiable algorithms analyze the solution optimality \emph{after} the optimization and provide a decisive metric on optimality. Considering this fact alone, certifiable algorithms do not inherently provide a solution to perceptual aliasing. On the other hand, robust norms filter noisy measurements to reveal the underlying optimization minima. However, in the absence of information or the presence of perceptual aliasing, robust norms do not help to overcome the underlying problem.

\subsection{Degeneracy Detection}\label{subsection:degeneracy_detection}
Point cloud registration techniques have demonstrated reliable performance in various practical applications; however, the underlying iterative minimization problem can become degenerate in challenging geometrically self-symmetric or featureless environments. This problem is often noticed in the form of \textit{LiDAR-slip} when traversing along the self-similar directions of the environment, e.g., narrow and long building corridors, severely degrading the robot pose estimation performance. Many techniques have been proposed to detect degenerate conditions by modeling it as part of the uncertainty or covariance of the pose estimation process~\cite{censi, brossard, CELLO3D, deepICP}. However, a unified uncertainty representation is often not tractable regarding sources of individual errors and tends to be over-optimistic~\cite{bonnabel2016covariance}. Motivated by this reasoning, direct degeneracy detection methods have been proposed.
 
\subsubsection{Geometric Methods}\label{classification:geometry_based}
Geometric approaches utilize the relationship between the registration cost function and the environment to analyze the quality of the pose estimation process. Among the first to investigate the geometric stability of the point-to-plane ICP, ~\cite{gelfand2003} proposed a sampling-based method to select the most valuable points in a scan to improve the conditioning of the optimization process. 
Building on this idea ~\cite{kwok2016improvements} proposed improvements such as the iterative center of mass calculation, rotation normalization, and cyclic point addition to improve the efficiency of the method. In a similar direction, \mbox{IMLS-SLAM}~\cite{deschaud2018imls} includes the contribution of a point to the matching procedure to ensure the observability of the optimization; however, the additional overhead makes the method unsuitable for real-time applications.
Similarly, to estimate the localizability of an environment using the eigenspace of the Hessian, authors in ~\cite{tunnelLocalizability} propose to measure the constraint strength of a point and surface-normal pair as the sensitivity of measurements w.r.t to the optimization states. Although this formulation is theoretically grounded and well-structured, the given parameterization does not use the full parametric range of the localizability analysis to distinguish the localizable and non-localizable directions due to alignment based normalization. Furthermore, it does not account for the scale difference between rotation and translation sub-spaces, which can be significant in practical applications. Crucially, this method requires a prior-built point cloud map of the environment.
The proposed localizability-awareness framework, \mbox{X-ICP}, also uses a geometry-based approach, similar to~\cite{tunnelLocalizability}, but does not require a prior map of the environment, the scale difference between rotation and translation is taken into account, and the localizability detection is more fine-grained (three levels vs. binary).

\subsubsection{Optimization-based Methods}
\label{classification:system_state_based}
For degeneracy detection, several approaches have proposed different metrics to quantify the state of the optimization.
The work in~\cite{gramian} proposes to use the observability Gramian as a measure of insufficient sensor measurements required to constrain the optimization. Similarly,~\cite{planeRank, LION, dareSLAM} proposed to use the condition number of the optimization Hessian as a single combined degeneracy metric for all $6$-DoF of the pose estimation optimization problem. With similar reasoning, \cite{fisher1, fisher2} proposed to utilize the determinant of the fisher information matrix as the degeneracy detection metric instead. Reasoning that a combined degeneracy metric is not representative for both translation and rotational sub-spaces,~\cite{rankDeficientSLAM} proposes to utilize relative condition number only for the translational sub-space to detect optimization degeneracy along each translational direction. In contrast, CompSLAM~\cite{compSLAM} uses the D-optimality criterion~\cite{dOpt} as a degeneracy detection metric to detect under-constrained environments for different modalities of a robust sensor fusion.
Although practical in nature, these approaches subject different degeneracy metrics to a threshold to identify degeneracy, which is not only heuristic in nature but difficult to generalize as these metrics depend on the environment's structure and the amount of information observed during an operation instance. Furthermore, the smooth transition of the optimization problem from degenerate to non-degenerate is difficult to capture with a binary degeneracy detection method. 

For degeneracy-aware LiDAR-based SLAM systems, the seminal work of~\cite{solRemap} proposes both a degeneracy detection metric called \textit{degeneracy factor} and a degeneracy mitigation method named \textit{solution remapping}.
The \textit{degeneracy factor} utilizes the minimum eigenvalue of the Hessian matrix of the optimization to detect the degeneracy, and the \textit{solution remapping} method utilizes the detected degeneracy to project the optimization solution only along the well-constrained directions. This work has been adopted by multiple LiDAR-based SLAM frameworks~\cite{compSLAM, M_LOAM,solRemap_kaess, ren2021towards, zhou2020lidar} and is considered state-of-the-art; however, certain aspects can limit its efficacy. \textit{i)} Being binary in nature, the method depends on the heuristic tuning of thresholds for operation in different environments~\cite{nubert2022learning}. \textit{ii)} As eigenvalues represent the scale of their respective eigenvectors, thresholds for translation and rotation cannot be represented by a singular value. \textit{iii)} As \textit{solution remapping} projects the solution along the well-constraint directions, it assumes that well-conditioned directions remain completely unaffected by the ill-conditioning. However, this assumption may be incorrect in severely degenerate environments since the optimization might diverge \textit{before} the \textit{solution remapping} method can project it. Echoing similar reasoning, ~\cite{hinduja2019degeneracy} proposes to improve \textit{solution remapping} by using the relative condition number of the optimization directions to set the eigenvalue threshold automatically. These state-of-the-art methods~\cite{solRemap, hinduja2019degeneracy} are widely adopted in the field of robotics and are used as baseline methods for comparison in this work. 

\subsubsection{Data-driven Methods}\label{classification:data_driven}
With the advent of learning-based methods, data-driven methods provide promising alternatives to perform degeneracy detection.
The work in~\cite{nobili1} formulates localizability as a function of overlap between scans and uses a support vector classifier to learn a risk metric for point cloud registration. Similarly, OverlapNet~\cite{overlapnet} showed the importance of using overlap between point cloud scans to identify the similarity of environments. In contrast, ~\cite{CELLO3D} proposes direct localizability quantification using a learned pose estimation uncertainty metric. Furthermore, combining covariance estimation and localizability detection, ~\cite{deepLocalizability} proposes a deep-learned entropy-based metric. Although successful, these methods rely on extensive ground truth data for learning and are unsuitable for real-time operation. To alleviate reliance on data, ~\cite{nubert2022learning} proposes to leverage simulation for training and only consider the current LiDAR scan to predict a $6$-DoF localizability metric. The authors use sparse 3D convolutions to show generalization across different sensors and environments through real-world experiments. However, this approach is limited to scan-to-scan point cloud registration.

\subsection{Constrained Optimization in Point Cloud Mapping}\label{subsection:constrainedOptInMapping}
Constrained optimization techniques are well-known in literature; however, their application to point cloud registration has only recently attracted more attention. Among the first, ~\cite{flory2009constrained} presented a constrained optimization method for penetration-free point cloud registration, improving the quality of pose estimation.
Similarly, ~\cite{constraintOptimization} uses non-linear equality constraints to reduce the linearization error of rotation estimation for point cloud registration. To improve robustness against sensor noise and correspondence outliers, ~\cite{sparseICP} proposes to use an augmented Lagrangian to solve a constrained optimization problem by adding each measurement as a separate constraint. 
A recent work~\cite{ct_icp} introduces the addition of soft constraints as costs to the ICP optimization to ensure trajectory continuity between different scans.
In contrast, ~\cite{teaser} formulates the point cloud registration problem as a constrained quadratic program to provide globally optimal point cloud registration results. These methods demonstrate improved robustness and accuracy for the global point cloud registration problem; however, they do not address the utility of constraints towards limiting the effect of degeneracy in the optimization.

Addressing this limitation, using the degeneracy detection formulation from~\cite{solRemap}, Hinduja et al.~\cite{hinduja2019degeneracy} proposed a partial factor formulation to incorporate the point cloud registration result into a pose graph formulation in a degeneracy-aware manner. The authors used a re-projection matrix as covariance to reflect the well-constraint directions of the registration result. While this method shows a straightforward way of integrating degeneracy awareness into sensor fusion frameworks, the adverse effects of degeneracy on low-level ICP optimization are not addressed. 

On the other hand, the authors in~\cite{rankDeficientSLAM} propose to use the relative condition number to detect degeneracy and to penalize the motion change along the degenerate directions by introducing constraints in a factor graph formulation in the form of a smooth cost function.
However, this work is only tested in 2D scenarios and does not consider introducing constraints directly in the optimization process. Given the discussion, the use of degeneracy analysis to constrain the ICP registration problem for robot operation in challenging and degenerate environments remains an open problem that this work aims to address.

\begin{figure*}[t]
\centering
\includegraphics[width=1.0\linewidth]{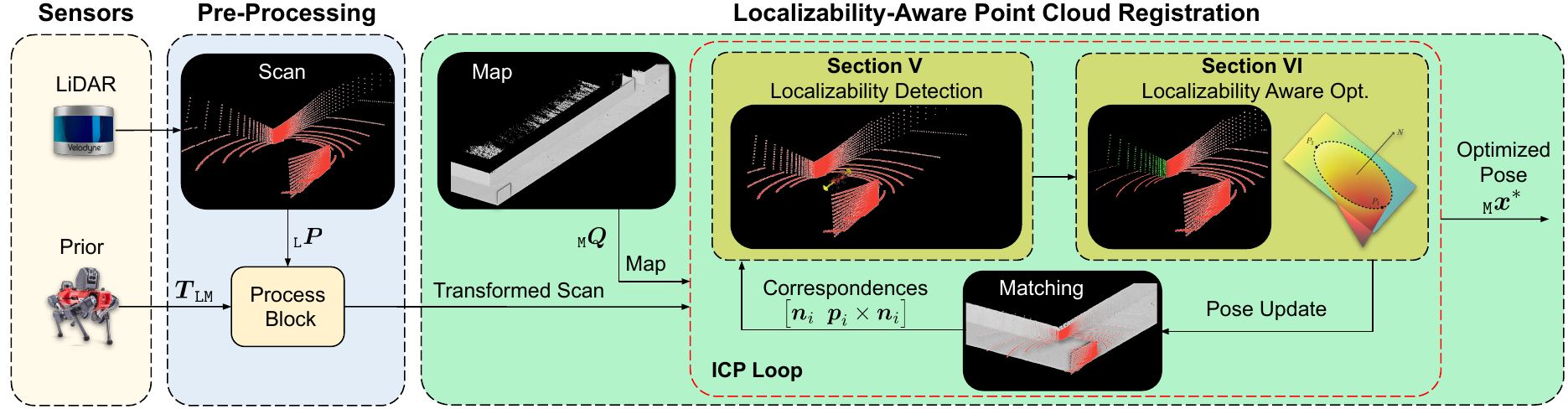}
\caption{Overview of the proposed localizability-aware point cloud registration framework. The pose prior is used to transform and undistort the input point cloud, which is, together with the existing point cloud map, fed to the iterative ICP optimization loop. The optimized drift-free pose within the ICP loop is calculated using the proposed localizability detection (Section~\ref{section:detection}) and aware optimization (Section~\ref{section:aware_opt}) modules.}
\label{fig:highLevelOverview}
\vspace{-1.0em}
\end{figure*}

\section{Problem Formulation \& Preliminaries}
\label{sec:problem_formulation}

This section provides an overview of the point cloud registration process and formulates the problem for operation in LiDAR degenerate environments. All vectors and matrices are denoted in \textbf{bold}, with matrices expressed with \textbf{C}apital letters.
\subsection{Point Cloud Registration}
\label{sec:problem_formulation_registration}
The problem of point cloud registration is defined as finding the rigid body transformation $\boldsymbol{T}_{\mathtt{M}\mathtt{L}} \in SE(3)$, that best aligns a \textit{reading} point cloud of $N_p$ points $_{\mathtt{L}}\boldsymbol{P} \in \mathbb{R}^{3 \times N_p}$ in LiDAR frame (denoted as $\mathtt{L}$) to a \textit{reference} point cloud of $N_q$ points,
$_{\mathtt{M}}\boldsymbol{Q} \in  \mathbb{R}^{3 \times N_q}$ in map frame (denoted as $\mathtt{M}$).
The rigid transformation $\boldsymbol{T}_{\mathtt{M}\mathtt{L}} = \big[\boldsymbol{R}_{\mathtt{M}\mathtt{L}}\:|\: _\mathtt{M}\boldsymbol{t}_{\mathtt{M}\mathtt{L}}\big]$, consists of a rotation matrix $\boldsymbol{R} \in  SO(3)$, and a translation vector $\boldsymbol{t} \in \mathbb{R}^3$, with $\boldsymbol{t} = [t_x,\: t_y,\: t_z]^{\top}$. 
For each reading point $_{\mathtt{L}}\boldsymbol{p} \in \mathbb{R}^3$ in $_{\mathtt{L}}\boldsymbol{P}$, the closest reference point $_{\mathtt{M}}\boldsymbol{q} \in \mathbb{R}^3$ in $_{\mathtt{M}}\boldsymbol{Q}$,  is found through a correspondence search, often in the form of a \textit{k-d tree} search. 
This data association process is defined as $\mathcal{M} \in \mathbb{R}^{6 \times N} = \text{matching}\big(\, _{\mathtt{L}}\boldsymbol{P},\ _{\mathtt{M}}\boldsymbol{Q},\ \boldsymbol{T}_{\mathtt{L} \mathtt{M}, \text{init}}\big) = \left\{ \big(\,_{\mathtt{M}}\boldsymbol{p}, \: \{_{\mathtt{M}}\boldsymbol{q}, \ _{\mathtt{M}}\boldsymbol{n}\}\big)\: :(_{\mathtt{M}}\boldsymbol{p} \in {_{\mathtt{M}}{\boldsymbol{P}}}), \big({_{\mathtt{M}}{\boldsymbol{q}}} \in {_{\mathtt{M}}{\boldsymbol{Q}}}\big) \right\}$ where $_{\mathtt{M}}\boldsymbol{p}$ and $ _{\mathtt{M}}\boldsymbol{q}$ are the matched point pairs and $_{\mathtt{M}}\boldsymbol{n} \in \mathbb{R}^3, \: \| {_{\mathtt{M}}\boldsymbol{n}} \| =1$ is the surface-normal vector of point $_{\mathtt{M}}\boldsymbol{q}$. Furthermore, $N \leq N_p$ is the number of matched points and indicates the size of the problem for the rest of the work. 
The initial transformation $\boldsymbol{T}_{\mathtt{L} \mathtt{M}, \text{init}}$ is provided as an initial guess to transform the scan data to the reference frame, to improve the matching process and optimization convergence characteristics. While the accuracy of this initial transformation is critical for the convergence of the minimization~\cite{brossard}, an analysis of the effect and quality of this initial transformation is outside the scope of this work.

Multiple error functions have been proposed for point cloud alignment; in this work, the point-to-plane~\cite{point_to_plane} cost function is used. The ICP minimization problem with the point-to-plane cost function can be defined as follows:
\begin{align}
\mathop{\text{min}}_{\boldsymbol{R},\: \boldsymbol{t}} 
\sum_{i=1}^N \Big|\Big| \big((\boldsymbol{R} \boldsymbol{p}_i + \boldsymbol{t}) - \boldsymbol{q}_i\big) \cdot {\boldsymbol{n}}_i \Big|\Big|_2.
 \label{eq:point_to_plane_cost}
\end{align}

Different solvers, such as singular value decomposition~(SVD)~\cite{svd}, LU decomposition, Gauss-Newton, and \mbox{Levenberg-Marquardt}, can be used to solve this minimization problem. In this work, the focus lies on direct linear algebra solvers like SVD, which exists for all matrices. 

Following the derivation of Pomerleau et al.~\cite{ICPderivation}, the minimization~\eqref{eq:point_to_plane_cost} can be reformulated as a quadratic cost optimization problem as:
\begin{align}\label{eq:huge_cost}
    \begin{split}
\mathop{\text{min}}_{\boldsymbol{x}\in \mathbb{R}^{6}} 
\boldsymbol{x}^T
\underbrace{
\Bigg(
  \sum_{i=1}^N
      \underbrace{\begin{bmatrix} \left(\boldsymbol{p}_i \times \boldsymbol{n}_i\right) \\  \boldsymbol{n}_i \end{bmatrix}}_{\boldsymbol{A}}
      \underbrace{\left[ (\boldsymbol{p}_i \times \boldsymbol{n}_i)^T \  \boldsymbol{n}_i^T \right]}_{\boldsymbol{A}^\top}
\Bigg)
}_{\boldsymbol{A}^\prime}
\boldsymbol{x}
\\ -
2\boldsymbol{x}^T
\underbrace{
\Bigg(
  \sum_{i=1}^N 
      \underbrace{\begin{bmatrix} \left(\boldsymbol{p}_i \times \boldsymbol{n}_i\right) \\  \boldsymbol{n}_i \end{bmatrix}}_{\boldsymbol{A}}
      \boldsymbol{n}_i^T
      (\boldsymbol{q}_i-\boldsymbol{p}_i)
\Bigg)
}_{\boldsymbol{b}^\prime} + \; \text{Const.}.
    \end{split}
\end{align}

Here $\boldsymbol{x} = [\boldsymbol{r}^\top \ \boldsymbol{t}^\top]^\top \: \in  \mathbb{R}^{6}$ are the optimization variables, with $\boldsymbol{r} \in \mathfrak{so}(3)$ being the rotation vector $\left(\text{lie algebra of}~SO(3) \right)$ and $\boldsymbol{t} \in \mathbb{R}^3$.
Moreover, $\boldsymbol{A}^\prime \in \mathbb{R}^{6\times 6}$ denotes the Hessian of the optimization problem, and $\boldsymbol{b}^\prime \in \mathbb{R}^6$ incorporates the constraints between the point clouds. The Hessian constitutes the second-moment matrix of the optimization and defines the local behavior of the Jacobian. Moreover, the Equation~\eqref{eq:huge_cost} with optimization variable $\boldsymbol{x}$ can be reformulated as a least squares optimization problem as follows:
\begin{align}
\mathop{\text{min}}_{\boldsymbol{x}\in \mathbb{R}^{6}} 
\Big|\Big|\boldsymbol{A}^\prime
\boldsymbol{x}- \boldsymbol{b}^\prime\Big|\Big|_2.
\label{eq:final_cost}
\end{align}

Solving this minimization problem is simple for a (semi-) positive definite matrix $\boldsymbol{A}^{\prime}$. The solution of this $6\times 6$ linear equation system will result in the optimal alignment translation vector $\boldsymbol{t}$ and rotation $\boldsymbol{r}$ under the performed linearizations. During ICP, due to non-linearities and the resulting iterative nature of the algorithm, the described operations are repeated until convergence.

\subsection{Operation in Degenerate Environments}
In practical applications, the described point cloud registration process can fail due to LiDAR degeneracy induced by the absence of geometrically informative structures. The solution $\boldsymbol{T}_{\mathtt{M}\mathtt{L}}$ of the registration step becomes under-constrained, which means that one or multiple dimensions of the $6$-DoF transformation are (almost) not observable from the point correspondences. 
Consequently, the primary focus of this work is to address the problem of finding the optimal transformation $\boldsymbol{T}_{\mathtt{M}\mathtt{L}}$ and determining the difficult-to-estimate directions in the presence of environmental degeneracy. While previous research on point cloud registration tends to disregard this particular scenario, this work introduces a dedicated solution to tackle these challenging situations, allowing for effective operation even in extreme scenarios.

\section{System Overview}
\label{section:System_Overview_and_problem_formulation}
To reliably perform point cloud registration \textit{and} pose estimation in LiDAR degenerate environments, this work proposes \mbox{X-ICP}, a localizability-guided constrained point cloud registration method. An overview of the proposed framework is shown in Fig.~\ref{fig:highLevelOverview}. 
The proposed components for detecting and mitigating LiDAR degeneracy are denoted as \textbf{Loc.-Module} and \textbf{Opt.-Module}. For demonstration purposes, both components are embedded into a scan-to-map ICP registration system~\cite{pharosHiltiReport} developed by ANYbotics, which is based on a modified version of \textit{libpointmatcher}, an open-sourced point cloud registration library~\cite{libpointmatcher}.
The mapping pipeline based on this point cloud registration framework runs at \SI{5}{\hertz} and uses the common \mbox{point-to-plane} ICP cost function. Note that the proposed contributions of this work are applicable to any point-to-plane iterative optimization-based registration framework and are independent of the scan-to-map implementation. As discussed in Section~\ref{sec:problem_formulation_registration}, the robustness and accuracy of the scan-to-map registration depend on the quality of the initial guess. Hence, \mbox{X-ICP} performs better with a robust state estimator compared to a dead-reckoning system.


\begin{figure*}[t]
    \centering
    \includegraphics[width=1.0\linewidth]{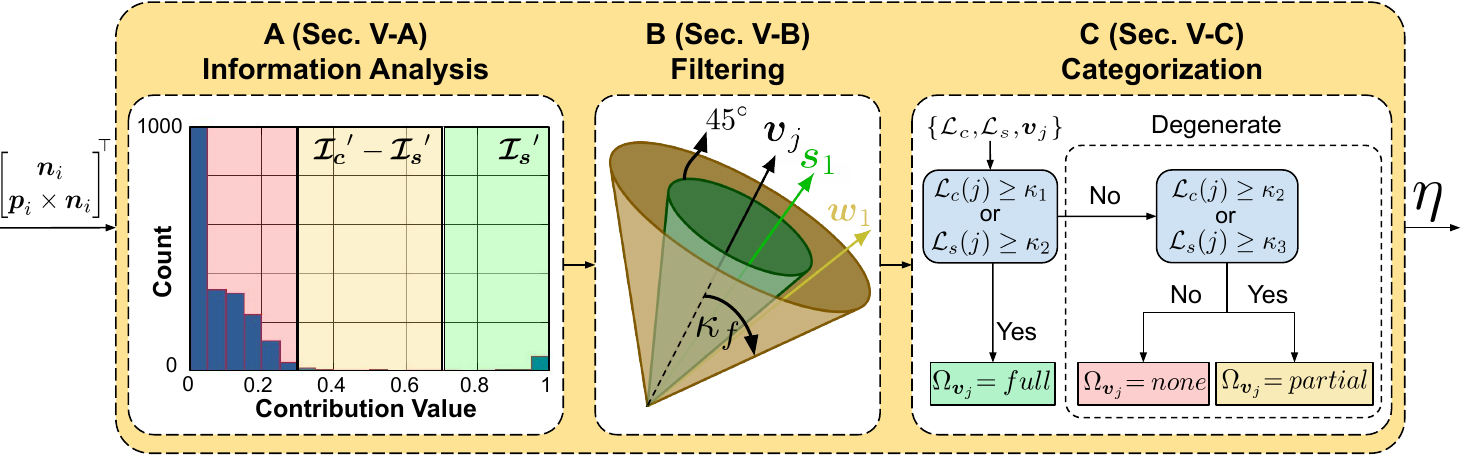}
    \caption{Overview of the localizability detection module. \textbf{A - Information Analysis:} An exemplary histogram shown for one eigenvector $\boldsymbol{v}_j$ direction. The histogram is color-coded to illustrate the strong- (\protect\greenline) and the weak-contribution (\protect\yellowline) regions. The red (\protect\redline) region contains the to-be-filtered information pairs. \textbf{B - Filtering:} This step measures the alignment of all vectors in the histogram to the eigenvector $\boldsymbol{v}_j$, with $(\boldsymbol{s}_1,\boldsymbol{w}_1)$ illustrating strong and weak alignment vectors, respectively. \textbf{C - Categorization:} Localizability categories are assigned to each optimization eigenvector based on a decision tree.}
\label{fig:localizabilityModuleOverview}
\vspace{-1.0em}
\end{figure*}

\subsection{Localizability Detection Module Overview} 
\label{section:localizability_module_overview}
The goal of the \textbf{Loc.-Module} is to approximate the null space of the ICP optimization Hessian $\boldsymbol{A}^\prime$. The \textit{localizability vector} $\eta$ (spanning all $6$-DoFs of the optimization problem) is introduced to achieve this goal. This localizability vector indicates which eigenvectors should be considered to lie within Hessian's null space. Since the localizability analysis is performed directly in eigenspace (using the eigenvectors), the degenerate direction of the environment does not need to align with either the robot or the map frame. The localizability vector is defined as 
\begin{align}
\begin{split}
\displaystyle\boldsymbol{\eta}&=\{\boldsymbol{\eta}_t,\;\boldsymbol{\eta}_r\}\in \mathbb{R}^{6} \\
&=\big(\{{\eta}_{_{\mathtt{L}}\boldsymbol{v}_{t_1}},
{\eta}_{_{\mathtt{L}}\boldsymbol{v}_{t_2}},
{\eta}_{_{\mathtt{L}}\boldsymbol{v}_{t_3}}\},
\{{\eta}_{_{\mathtt{L}}\boldsymbol{v}_{r_1}},
{\eta}_{_{\mathtt{L}}\boldsymbol{v}_{r_2}},
{\eta}_{_{\mathtt{L}}\boldsymbol{v}_{t_3}}\}\big)^{\top}, 
\end{split}
\label{eq:localizability_categories}
\end{align}
where ${_{\mathtt{L}}{\boldsymbol{v}_{t_i}}} = \boldsymbol{V}_t(\dots, \ j),\: \forall~j \in \{1,\ 2,\ 3\}$ are the translational eigenvectors of the Hessian $\boldsymbol{A}^\prime$ corresponding to $\boldsymbol{t}$ variables and expressed in the LiDAR frame, whereas ${_{\mathtt{L}}{\boldsymbol{v}_{r_i}}} = \boldsymbol{V}_r(\dots, \ j),\: \forall~j \in \{1,\ 2,\ 3\}$
correspond only to the rotation $\boldsymbol{r}$. More details on how to obtain the eigenvector matrices $\boldsymbol{V}_t$ and $\boldsymbol{V}_r$ are explained in Section~\ref{task:Information_Analysis}.
The localizability vector $\boldsymbol{\eta}$ represents the localizability state of each eigenvector in the form of a categorical variable; these discrete set of categorical variables are defined as $\boldsymbol{\eta}_{j} \in \{none,\ partial,\ full\}$ with $j \in \{1, \cdots 6\}$, where the categories correspond to being non-localizable, partially-localizable, and localizable, respectively. The sequence of actions for each localizability category will be explained in Section~~\ref{task:categorization}.

\subsection{Localizability Aware Optimization Module Overview} 
Utilizing the output of the Loc.-Module, i.e., the discrete set of localizability categories $\boldsymbol{\eta}$, the goal of the \textbf{Opt.-Module} is constructing and solving the constrained optimization problem to estimate the optimal state $\boldsymbol{x}^{*}$ for the optimization problem~\eqref{eq:final_cost}, and is explained in Section~\ref{section:aware_opt}. 
In this part, constrained optimization techniques based on Lagrangian-multipliers are used to obtain the best possible solution given the observed localizability. The possible outcomes include the registration initial guess to be unchanged in the directions where \mbox{$\boldsymbol{\eta}_{j} = none$}, changed in a controlled manner for \mbox{$\boldsymbol{\eta}_{j} = partial$} or updated without any constraints for \mbox{$\boldsymbol{\eta}_{j} = full$}.

The complete \mbox{X-ICP} framework (cf. Fig.~\ref{fig:highLevelOverview}) offers reliable robot pose estimation in the presence of LiDAR degeneracy. The main contributions of this work are explained and highlighted in Sections~\ref{section:detection} and \ref{section:aware_opt}.

\section{Localizability Detection Module}
\label{section:detection}
In this section, the details of the \textbf{Loc.-Module} are described according to the information flow shown in Fig.~\ref{fig:localizabilityModuleOverview}. Localizability detection aims to measure the information within the correspondences to identify the under-constrained directions correctly. 
As shown in Fig.~\ref{fig:localizabilityModuleOverview}, given the correspondences, the first task is the \textbf{Information Analysis} indicated by Fig.~\ref{fig:localizabilityModuleOverview}-A, analyzing the relations between the Hessian and the geometric information from the environment. 
In the second step, the \textbf{Filtering} (Fig.~\ref{fig:localizabilityModuleOverview}-B), the redundant contribution values from the {Information analysis} (Fig.~\ref{fig:localizabilityModuleOverview}-A) step are filtered out. 
 Finally, the filtered contribution information is inferred, leading to a fine-grained \textbf{Categorization} step, as shown in Fig.~\ref{fig:localizabilityModuleOverview}-C. 
After the correspondence search, the matched correspondences are transformed back to the LiDAR frame ($\mathtt{L}$) to eliminate the scale effect of the map's physical size on the localizability of rotational eigenvectors, and the localizability analysis is performed in this frame. 

\subsection{Information Analysis}\label{task:Information_Analysis}
\subsubsection{Principal Component Analysis}
The information analysis starts with an eigenvalue analysis of the Hessian matrix of the optimization problem. For the derivation of the matrix for a point-to-plane ICP cost function, the reader is referred to Section~\ref{sec:problem_formulation}, where the Hessian is provided as $\boldsymbol{A^\prime}$ in Equation~\eqref{eq:final_cost}. 
The Hessian can be divided into sub-matrices based on the relation to the minimization variables $\boldsymbol{x}$:
\begin{align*}
{\boldsymbol{A}}^{\prime} = 
\begin{bmatrix}
{\boldsymbol{A}}^{\prime}_{rr}  & {\boldsymbol{A}}^{\prime}_{rt}      \\
{\boldsymbol{A}}^{\prime}_{tr}  & {\boldsymbol{A}}^{\prime}_{tt}        \\
            \end{bmatrix}_{6\times6}. 
\end{align*}
Here, ${\boldsymbol{A}}^{\prime}_{rr} \in \mathbb{R}^{3\times 3}$ exclusively contains information related to the rotation variables. Similarly, ${\boldsymbol{A}}^{\prime}_{tt} \in \mathbb{R}^{3\times 3}$ exclusively contains information related to the translation variables. 

Moreover, as discussed in Section~\ref{sec:problem_formulation_registration}, the ICP Jacobian consists of two independent elements, the $\boldsymbol{n}$ and $\boldsymbol{p}\times\boldsymbol{n}$ for translation and rotation, respectively. 

As it is not trivial to treat $t \in \mathbb{R}^3$ and $r \in \mathfrak{so}(3)$ together due to differences in scale and type, only
${\boldsymbol{A}}^{\prime}_{tt}$ and ${\boldsymbol{A}}^{\prime}_{rr}$ are included in the eigen-analysis of the SVD. If all the elements of the Hessian are used for the eigen-analysis, the scale difference between rotation and translation elements will complicate setting heuristic free localizability parameters. For the rotation and translation components, the resulting eigen-decomposition is given as
\begin{align*}
{\boldsymbol{A}}^{\prime}_{tt} = \boldsymbol{V}_{t}\Sigma_t \boldsymbol{V}_{t}^{\top}, \quad     {\boldsymbol{A}}^{\prime}_{rr} = \boldsymbol{V}_{r}\Sigma_r\mathbf{V}_{r}^{\top},
\end{align*}
where $\boldsymbol{V}_t \in \mathbb{R}^{3\times3}$ and $\boldsymbol{V}_r \in \mathbb{R}^{3\times3}$ are the eigenvectors in matrix form, 
and $\Sigma_t \in \{\text{diag}(\boldsymbol{v}): \boldsymbol{v} \in \mathbb{R}^n_{\geq 0} \}$ and $\Sigma_r \in \{\text{diag}(\boldsymbol{v}): \boldsymbol{v} \in \mathbb{R}^n_{\geq 0} \}$ are diagonal matrices with the eigenvalues of $\boldsymbol{A}^{\prime}_{tt}$ and $\boldsymbol{A}^{\prime}_{rr}$ as the diagonal entries, respectively.
Intuitively, the eigenvalues in $\Sigma_t$ and $\Sigma_r$ provide a direct measure of the information along each eigenvector it is paired with. However, as discussed in Section~\ref{section:related_works}, eigenvalues can behave inconsistently in different environments and for varying sensors, and hence are not used directly as part of this work's localizability estimation.

\subsubsection{Information Pair Contribution}
The second part of the proposed information analysis is to formulate the contribution of each \textbf{information pair}, defined as $\big(_{\mathtt{L}}{\boldsymbol{p}},\; _{\mathtt{L}}{\boldsymbol{n}}\big)$. A formal relationship is required between the information pairs and the optimization cost to assess how much a pair contributes to the cost. Gelfand et al.~\cite{gelfand2003} showed that the squared summation of these contributions provides reasonable estimates of the eigenvalues of the optimization Hessian, suggesting that the Jacobian-based localizability formulation and eigenvalues are correlated. Similar to other works~\cite{gelfand2003, kwok2016improvements, DNSS}, \mbox{X-ICP} utilizes the elements of the \mbox{Jacobian} (instead of the Hessian) directly, as inferred from Equation~\eqref{eq:huge_cost}, simplifying the formulation and allowing for more practical deployment in various environments while keeping the formulation correlated to the optimization Jacobian.

\begin{figure}[t]
\vspace{-1.0em}
\centering
\includegraphics[width=0.95\linewidth]{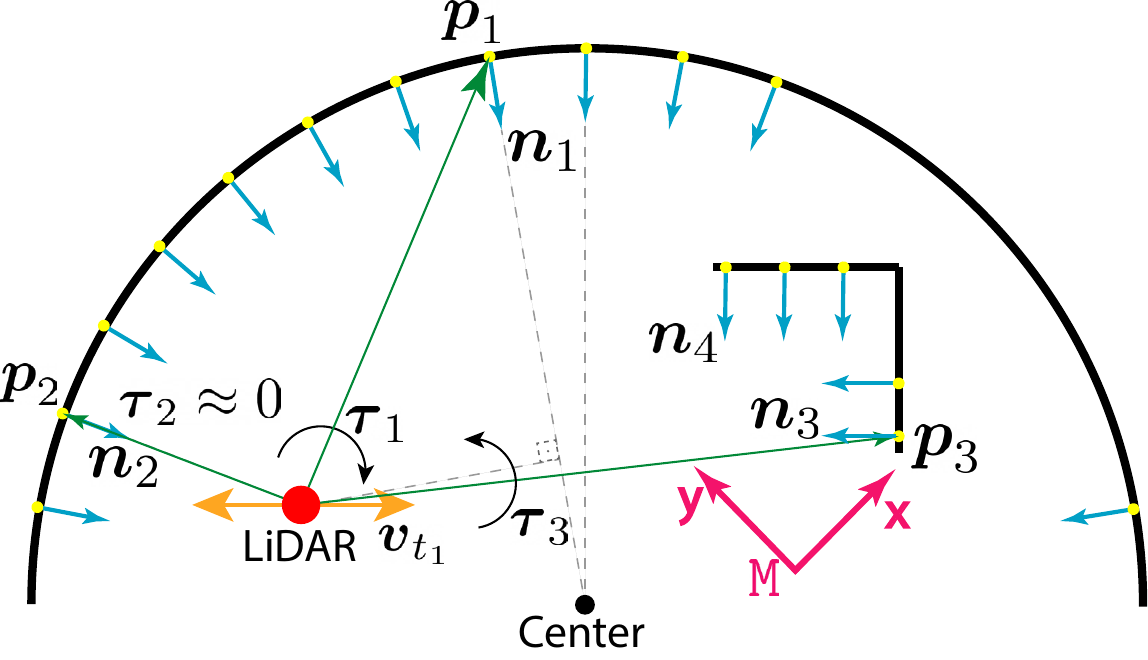}
\caption{A 2D-example illustrating the contribution to localizability. Points $\boldsymbol{p}$ (green arrows), surface normals $\boldsymbol{n}$ (blue arrows), and the LiDAR center (red dot \protect\redcircle) are shown. The span of one of the eigenvectors is depicted in orange. Finally, the induced torques ($\boldsymbol{\tau}$) are shown for three point- and normal-pairs.}
\label{fig:intution}
\vspace{-1.0em}
\end{figure}

\paragraph{Analogy to Classical Mechanics}
In an analogy to classical mechanics, the optimization \mbox{Jacobian} measures the magnitude of the wrench induced locally by each information pair. A wrench system consist of elements such as $\textbf{force}(\boldsymbol{n})$ and $\textbf{torque}(\boldsymbol{\tau} = \boldsymbol{p}\times\boldsymbol{n})$. An illustration of the underlying wrench system and localizability concept in 2D is provided in Fig.~\ref{fig:intution}, illustrating an environment consisting of two perpendicular and a semi-circular wall. Moreover, the shown map frame $\mathtt{M}$ does not need to align with the principal directions of the environment (e.g., $\boldsymbol{v}_{t_1}$), as the localizability detection of \mbox{X-ICP} is done in the optimization's eigenspace, rendering it invariant to the orientation of the robot with respect to the environment. 

Information analysis of the point and surface normal pairs $(\boldsymbol{p}_i,\boldsymbol{n}_i), \forall~i \in \{1,\, 2, \  3\}$ show different contributions towards the localizability of different (optimization) principal directions. As an example, the surface normals $\boldsymbol{n}_{1,2,3}$ provide translational localizability contributions towards the direction of the eigenvector $\boldsymbol{v}_{t_1}$(due to $\boldsymbol{n}_{1,2,3} \cdot \boldsymbol{v}_{t_1} > 0$), whereas the surface normal $\boldsymbol{n}_4$ does not provide any contribution for this direction. As another example, for the rotational contribution around the z-axis (in $\mathtt{M}$), it holds: $\tau_1 > \tau_3 > \tau_2 \approx 0$. 

Although intuitive, the torque formulation does not provide a generalizable localizability parameterization without normalization. The reason is that the point $\boldsymbol{p}$ can be at arbitrary long distances, leading to higher torque values, which limits the generalizability of localizability formulation in different environments for practical applications.
Kwok and Tang~\cite{kwok2016improvements} studied the influence of this scale difference for different normalization techniques in ICP. They suggest that \textit{maximum norm}-normalization performs better than \textit{moment}-normalization, or \textit{average norm}-normalization, as done in~\cite{gelfand2003}. 
This holds for obtaining a reliable solution to the ICP problem, where spatial relations between correspondence are crucial for the estimation. However, this is not necessarily the case for point-wise contribution calculation for localizability estimation. Considering this and the requirement that the point norm should not affect the contribution value directly, this work proposes to use moment normalization, which maps the torque values to a unit sphere (cf. Equation~\eqref{eq:stackedContributions}).

After the moment normalization of the individual torque vectors, the wrench system is stacked for all available information pairs to form the information matrices as follows:
\begin{align}
\begin{split}
\boldsymbol{\mathcal{F}}_r&=
\begin{bmatrix}
 \displaystyle\frac{\boldsymbol{p}_1\times\boldsymbol{n}_1}{\|\boldsymbol{p}_1\times\boldsymbol{n}_1\|_2}      &   \dots       & \displaystyle\frac{\boldsymbol{p}_N\times\boldsymbol{n}_N}{\|\boldsymbol{p}_N\times\boldsymbol{n}_N\|_2}      \\
\end{bmatrix}^{\top}, \\
\boldsymbol{\mathcal{F}}_t&=
\begin{bmatrix}
\boldsymbol{n}_1     &   \dots       & \boldsymbol{n}_N      \\
\end{bmatrix}^{\top}.
\end{split}
\label{eq:stackedContributions}
\end{align}
Here, $\boldsymbol{\mathcal{F}}_r \in \mathbb{R}^{{N\times3}}$ and $\boldsymbol{\mathcal{F}}_t \in \mathbb{R}^{{N\times3}}$ are the rotational and translational information matrices, respectively.
The final task in the information analysis step is to compute the localizability contribution from the information matrices $\boldsymbol{\mathcal{F}}_r$ and $\boldsymbol{\mathcal{F}}_t$. Instead of defining the localizability information in Equation~\eqref{eq:stackedContributions} as a function of quadratic terms (i.e \mbox{$\big(\boldsymbol{p} \times \boldsymbol{n}\big)^2$} and $\left(\boldsymbol{n}\right)^2$), it is defined linearly. The used terms \mbox{$\left(\boldsymbol{p} \times \boldsymbol{n}\right)$ and $ \left(\boldsymbol{n}\right)$} prevent scale compression of the contributions.

For numeric stability, if $\lvert\boldsymbol{\tau}\rvert$ is close to zero (i.e., the surface-normal and point vectors are near-parallel), the information pair is dropped. An example of this phenomenon is shown for the $\boldsymbol{p}_2,\:\boldsymbol{n}_2$ information pair depicted in Fig.~\ref{fig:intution}. Furthermore, the moment normalization is only applied for information pairs with $\lvert \boldsymbol{\tau} \rvert \geq 1$, preventing the inflation of torque values from within onto the unit sphere, which might push the Loc.-Module towards optimistic localizability estimation.

\paragraph{Localizability Contributions}
The localizability concept is defined in the eigenspace of the optimization, and hence, can be obtained for every eigenvector direction. 
This ensures that the detection is not affected by the orientation of the LiDAR or the robot w.r.t the environment, which is useful for practical applications. Thus, the information matrices defined in Equation~\eqref{eq:stackedContributions} are projected onto the eigenspace of the translation and rotation Hessians. To achieve this, the eigenvector matrices $\mathbf{V}_r$ and $\mathbf{V}_t$ will be used as follows:
\begin{align}
\boldsymbol{\mathcal{I}}_r = \big(\boldsymbol{\mathcal{F}}_r\cdot\mathbf{V}_{r}\big)^{|\cdot|}, \quad \boldsymbol{\mathcal{I}}_t = \big(\boldsymbol{\mathcal{F}}_t\cdot\mathbf{V}_{t}\big)^{|\cdot|}.
\label{eq:contributionsLoc}
\end{align}
Here, $\boldsymbol{\mathcal{I}}_r,\; \boldsymbol{\mathcal{I}}_t \in \mathbb{R}^{{N\times3}} $ are the localizability contributions for all information pairs $\{\boldsymbol{p},\;\boldsymbol{n}\}$, projected by the eigenvectors in $\{ \mathbf{V}_r,\;\mathbf{V}_t\}$.  
The $(\dots)^{|\cdot|}$ operator indicates the element-wise absolute value of the vector. Concurrently, the scalar values in $\boldsymbol{\mathcal{I}}_r$ and $\boldsymbol{\mathcal{I}}_t$ are direct indicators of localizability contribution of a certain direction. If the scalar value is $\mathcal{I}(i, \ \dots)=1.0, i \in {1,\dots,N}$, the direction's localizability contribution is maximum, and if $\mathcal{I}(i, \ \dots)=0.0, \ i \in {1,\dots,N}$, the information pair has no contribution to the localizability of the eigenvector direction.

An example of the localizability contributions is provided with a histogram as shown in \mbox{Fig.~\ref{fig:localizabilityModuleOverview}-A} for a single translation eigenvector. The contribution values are dominantly at the lower-end, indicating low contribution. Nevertheless, there is a high peak at around the contribution value of $1.0$, suggesting the presence of a small but highly contributing structure. While pair-wise quantification of localizability contributions is essential, this pair-wise information needs to be filtered and consolidated for categorizing the localizability state of the eigenvectors of the optimization Hessian.

\subsection{Filtering}\label{task:Filtering}
Given the localizability contributions defined in Equation~\eqref{eq:contributionsLoc}, $\boldsymbol{\mathcal{I}}=[\boldsymbol{\mathcal{I}}_r \:,\: \boldsymbol{\mathcal{I}}_t] \in \mathbb{R}^{N\times6}$, the goal of the \textbf{filtering} step is to remove redundant information and to render the present information interpretable. 

\subsubsection{Filtering Low Contribution}
\label{sec:low_contrib_filtering}
An example of redundant information is shown in Fig.~\ref{fig:localizabilityModuleOverview}-A, where the low contribution region highlighted in red (\protect\redline) dominates the analysis. If the localizability contribution is small, it might become indistinguishable from measurement- or feature extraction noise. This step addresses this issue by employing binary element-wise filtering as an outlier rejection step. The filtering operation is implemented as re-assignment:
\begin{align}
\boldsymbol{\mathcal{I}}^{\prime}_c(i,j)=
\begin{cases}
\boldsymbol{\mathcal{I}}(i,j),& \text{if } \boldsymbol{\mathcal{I}}(i,j)\geq \kappa_f\\
0,              & \text{otherwise}
\end{cases}.
\label{eq:corrected_localizability_contribution}
\end{align}
Here the indices are defined as $i \in \{1, \dots, N\}$ and $j \in \{1,\dots,6\}$. $\boldsymbol{\mathcal{I}_c}^{\prime}$ is the \textbf{filtered localizability contribution matrix}, which contains all reliable localizability contribution values. Moreover, $\kappa_f$ is the filtering parameter, the first user-defined parameter. Since this parameter captures the sensor and feature extraction noise for different \mbox{LiDAR} sensors, it should be re-adjusted. This parameter is set to \mbox{$\kappa_f = \text{cos}(80^\circ) \approx 0.1736$} throughout all Velodyne experiments. For the Seem\"uhle experiment with Ouster \mbox{OS0-128}, $\kappa_f$ is set to $cos(60^\circ) = 0.5$ due to a higher point variance noise-characteristic compared to the Velodyne \mbox{VLP-16}, resulting in more aggressive filtering; localizability contribution values with an alignment angle higher than $60^\circ$ are filtered out.

\subsubsection{Filtering High Contribution} 
Using this filtered localizability contribution $\boldsymbol{\mathcal{I}_c}^{\prime}$, the contributions larger than $\kappa_f$, can be combined to summarize the available geometrical information:
\begin{align}
\boldsymbol{\boldsymbol{\mathcal{L}}}_c(j)&= \sum_{i=1}^{N} \boldsymbol{\mathcal{I}_c}^{\prime}(i,j).
\end{align}
Here, $\boldsymbol{\mathcal{L}}_c \in \mathbb{R}^{1\times 6} $ is the \textbf{combined localizability contribution vector} over all reliable information pairs. Here, a higher number of filtered contribution values indicate higher available contribution information.
Matrix $\boldsymbol{\mathcal{I}_c}^{\prime}$ still contains weak but reliable contributions, which might be required during partial localizability. However, a measure of the strongest contributions is still required to identify the more fine-grained status of the localizability. 
The filtering step is based on \textbf{geometrical vector alignment}; only the alignment values greater than $\text{cos}(45^\circ) \approx 0.707$ will be considered a strong contribution, justified through geometric relations. In Fig.~\ref{fig:localizabilityModuleOverview}-B, the separation of strong and weak alignment regions against an eigenvector is visualized. The inner green cone indicates the region with strong vector alignment, whereas the yellow region indicates the weaker one. This separation is formulated as follows:
\begin{align}
\begin{split}
\boldsymbol{\mathcal{I}}^{\prime}_s(i,j)&=
\begin{cases}
\boldsymbol{\mathcal{I}}^{\prime}_c(i,j),& \text{if } \boldsymbol{\mathcal{I}}^{\prime}_c(i,j)\geq cos(45^\circ)\\
0,              & \text{otherwise},
\end{cases} \\
\boldsymbol{\mathcal{L}}_s(j)&= \sum_{i=1}^{N} \boldsymbol{\mathcal{I}}^{\prime}_s(i,j).
\end{split}
\end{align}
Here, $\boldsymbol{\mathcal{L}}_s \in \mathbb{R}^{1\times 6}$ is the \textbf{strong localizability contribution vector}. Similar to the combined localizability contribution, this expression is also affected by an increased number of filtered localizability contribution values; however, it is less sensitive to the sensor noise and only affected by incorrect point correspondence errors due to the second filtering step. These strong and combined localizability contribution vectors are crucial for the next step, where these vectors will be used to categorize the localizability.

\subsection{Categorization}\label{task:categorization}
In \mbox{X-ICP}, the localizability categorization is defined by a set of discrete categorical variables, as described in Section~\ref{section:localizability_module_overview}. In principle, the chosen localizability contribution parameterization allows for a continuous localizability formulation.
This continuous formulation could be included in the optimization objective as soft constraints in the form of additional cost elements in the optimization objective. However, removing the discretization step, introduces further challenges, such as balancing the cost elements and possible constraint violations.

Considering this point, \mbox{X-ICP} introduces discrete localizability categorization while keeping the underlying localizability contribution formulation continuous.
This discretization allows for the introduction of hard constraints into the least squares optimization problem in \mbox{X-ICP}, which is advantageous in degeneracy mitigation; the satisfaction of these constraints avoids divergence along the constrained directions.
\paragraph{Localizability Parameters}
To achieve this goal, three localizability parameters are introduced: \textit{i}) Parameter $\kappa_1$ is the safety localizability threshold defining the lower bound of being fully localizable. \textit{ii}) Parameter $\kappa_2$ regulates the transition from being localizable to partially-localizable and represents the upper bound of being partially localizable. \textit{iii}) Parameter $\kappa_3$ is the minimum information threshold that covers the case when the environment has sparse yet well-contributing information and regulates the transition from being partially-localizable to non-localizable.

\paragraph{Parameter Choice}
The three thresholds provide a natural relationship among each other: \mbox{$\kappa_1 \geq \kappa_2 > \kappa_3$}. These parameters are set based on the basin of convergence of the employed ICP algorithm. The rules for setting the parameters are as follows: 
\begin{enumerate}
\item As $\kappa_1$ sets the boundary between being localizable and \mbox{\{partial-localizability, non-localizable\}}, it can be set (almost) arbitrarily high. Note that a too high value of $\kappa_1$ introduces the additional computational cost of the localizability detection, while the partial-localizability utilizes the available information still adequately. An example value of $\kappa_1=250$ allows the ICP optimization to run with a minimum of $250$ perfectly aligned pairs. 
\item $\kappa_2$ is a parameter selected based on the system characteristics as robustness, and optimizer. Jointly with $\kappa_3$, it defines how the partial-localizability should be approached. It should be set between $\kappa_1$ and $\kappa_3$. 
\item Finally, $\kappa_3$ sets the boundary between the partial-localizable and non-localizable status. A value of $35$ allows the constrained ICP optimization to run in a controlled manner with a minimum of $35$ sampled pairs.
\end{enumerate}
Increasing $\kappa_1$ would make the degeneracy detection more aggressive, with only minimal added computational cost. On the other hand, $\kappa_2$ sets the \emph{uncertain} region, for which partial-localizability takes place. This allows the user to adjust the behavior of the degeneracy awareness framework in exchange for the computational cost of correspondence re-sampling. Lastly, lowering $\kappa_3$ increases the localizability risk of the system, which is user-defined.
Using these intuitions and the noise properties of point-to-plane ICP, in this work, the localizability parameters are set to $\kappa_1=250$, $\kappa_2=180$, and $\kappa_3=35$ \textbf{for all environments and sensors} shown in Section~\ref{section:results}.

\paragraph{Decision Tree}
These parameters are used in a decision tree to get the localizability categories as shown in \mbox{Fig.~\ref{fig:localizabilityModuleOverview}-C}. The decision tree takes the filtered localizability contribution matrices and an eigenvector as input. The decision tree operates eigenvector-wise; thus, the required binary comparisons are repeated for all $6$ principal directions. First, for each eigenvector direction, $\kappa_1$ is compared against $\boldsymbol{\mathcal{L}}_c$, and $\kappa_2$ against $\boldsymbol{\mathcal{L}}_s$. If either of these comparisons suggests that the problem is well-constrained, then the direction is localizable, $\eta_{\boldsymbol{v}_j}=full$. Secondly, suppose the first comparisons suggest that the problem is not well-constrained; in this case, $\kappa_2$ is compared against $\boldsymbol{\mathcal{L}}_c$ and $\kappa_3$ against $\boldsymbol{\mathcal{L}}_s$ to understand the amount of present information. If either of these comparisons holds, the localizability category is assigned as partial, \mbox{$\eta_{\boldsymbol{v}_j}=partial$}, otherwise it is assigned non-localizable, \mbox{$\eta_{\boldsymbol{v}_j}=none$}.

\begin{figure}[t]
    \centering
    \includegraphics[width=1.0\linewidth]{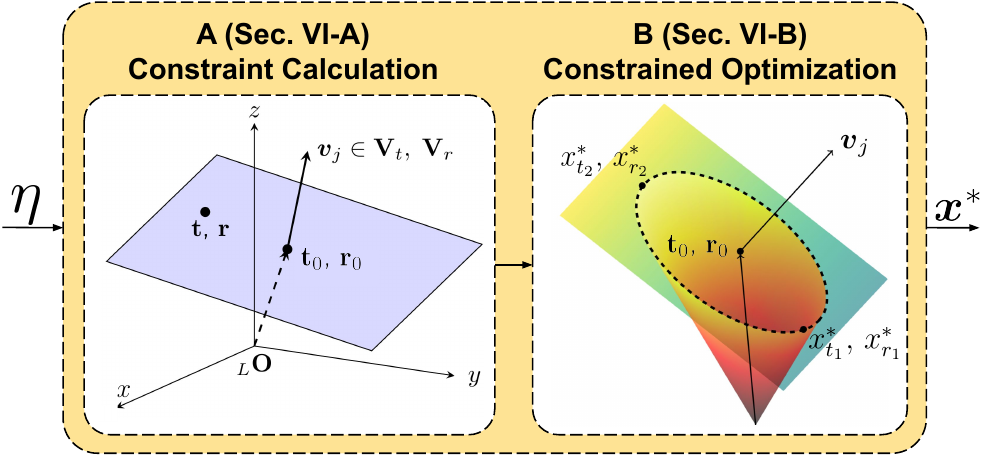}
     \caption{Overview of the constrained optimization module. The Opt.-Module steps are: \textit{i)} linear \textbf{Constraint Calculation}, which is in the form of a 3D plane, and \textit{ii)} \textbf{Constrained Optimization} which employs these constraints.}
    \label{fig:optimizationModuleOverview}
\end{figure}

\section{Localizability Aware Optimization Module}\label{section:aware_opt}
The \textbf{Opt.-Module} is responsible for the calculation and usage of additional constraints to perform reliable optimization in cases of degeneracy. An overview of this module is presented in Fig.~\ref{fig:optimizationModuleOverview}. Given the localizability categories $\boldsymbol{\eta}$ and eigenvectors $_{\mathtt{L}}\boldsymbol{v}_j$, the objective of this module is to find the optimal solution $\boldsymbol{x}^{*}$, even for an ill-conditioned optimization. 
In Section~\ref{task:contraint_calculation}, the calculation of optimization constraints for each localizability category is detailed and referred to as \textit{Constraint Calculation}. 
In Section~\ref{task:contrained_optimization}, these constraints are then integrated into the optimization problem (\textit{Constrained Optimization}). Since the full point cloud registration is performed in the map frame $\mathtt{M}$, all data in this section is also expressed in the map frame. At each iteration of ICP, also the eigenvectors are rotated to the map frame by applying \mbox{$\boldsymbol{R}_{\mathtt{M},\mathtt{L}_k} = \boldsymbol{R}_{\mathtt{M},\mathtt{L}_1} \cdot \boldsymbol{R}_{\mathtt{L}_1,\mathtt{L}_2} \cdot \dots \cdot \boldsymbol{R}_{\mathtt{L}_{k-1},\mathtt{L}_k} $}, 
where $k$ refers to the current index of the ICP iteration.

\subsection{Constraint Calculation}
\label{task:contraint_calculation}
There are multiple ways to enforce linear equality constraints during optimization. In this module, the constraint should limit the solution space of the optimization in 3D as depicted in Fig.~\ref{fig:optimizationModuleOverview}-A, along the ill-conditioned directions. Corresponding optimization eigenvector directions that are going to be constrained are expressed in the same geometrical frames as the optimization variables $\boldsymbol{t}$ and $\boldsymbol{r}$. Each constraint direction is given by $\boldsymbol{v}_{\{t,r\}_j}, \ j \in \{1,2,3\}$, and the admissible magnitude of the pose update is defined by values $\boldsymbol{t}_0$ and $\boldsymbol{r}_0$ which are calculated per constraint direction. The constraints are defined as
\begin{align}
\begin{split}
    \boldsymbol{v}^{\top}_{t_j}\cdot (\boldsymbol{t}-\boldsymbol{t}_{0})=0,\\
    \boldsymbol{v}^{\top}_{r_j}\cdot (\boldsymbol{r}-\boldsymbol{r}_{0})=0,
\end{split}
 \label{eq:constraint_form}
\end{align}  
where $\boldsymbol{v}^{\top}_{\{t,r\}_j}$ are eigenvectors in translational ($\in \boldsymbol{V}_t$) and rotational subspace ($\in \boldsymbol{V}_r$), respectively. Each line of Equation~\eqref{eq:constraint_form} defines a 3D plane with normal vectors $\boldsymbol{v}_{\{t,r\}_j}$ and plane points $\boldsymbol{t}_0$ and $\boldsymbol{r}_0$. 

Conceptually, $\boldsymbol{t}_0$ and $\boldsymbol{r}_0$ represent the pose estimates in the direction of the eigenvector they correspond to. An example of such a single constraint is illustrated in Fig.~\ref{fig:optimizationModuleOverview}-A. Given this constraint, a certain amount of the ICP pose update is imposed by $\boldsymbol{t}_0$ or $\boldsymbol{r}_0$, while the final optimized solution ($\boldsymbol{x^*_t},\boldsymbol{r^*_r}$) has to lie on the constraint plane as shown in Fig.~\ref{fig:optimizationModuleOverview}-B.

\subsubsection{Localizability Categories}
Optimization constraints are added based on the localizability category of the corresponding eigenvector $\boldsymbol{v}_j$. Here, the implications of different types of localizability categories are explained:
\begin{itemize}
    \item \scalebox{1.25}{$\eta_{\boldsymbol{v}_j}\hspace{-1mm}=$} $\boldsymbol{full}$: The optimization is carried out nominally without added constraints.
    \item \scalebox{1.25}{$\eta_{\boldsymbol{v}_j}\hspace{-1mm}=$} $\boldsymbol{none}$: The optimization direction along $\boldsymbol{v}_j$ is found to be non-localizable, and a constraint in this direction is employed by setting $\boldsymbol{t}_0$ or $\boldsymbol{r}_0$ to $\boldsymbol{0}_{3\times1}$. The pose update calculated by the ICP optimization will vanish along directions $\eta_{\boldsymbol{v}_j}=none$.
    \item \scalebox{1.25}{$\eta_{\boldsymbol{v}_j}\hspace{-1mm}=$} $\boldsymbol{partial}$: The optimization direction along $\boldsymbol{v}_j$ is found to be partially-localizable. The partial constraint value is calculated by re-sampling the available information from the correspondences, limiting the pose update in the direction of $\boldsymbol{v}_j$. The exact procedure is explained in the following section.
\end{itemize}

\subsubsection{Partial Localizability}
Calculating the \textit{partial} localizability constraint values requires re-sampling the ICP correspondences $\mathcal{M}$. The number of pairs that need to be re-sampled depends on the categorization process (cf. \mbox{Fig.~\ref{fig:localizabilityModuleOverview}-C}). 
If the statement $\boldsymbol{\mathcal{L}}_c \geq \kappa_2$ is true, then the information pairs used for calculating $\boldsymbol{\mathcal{L}}_c$ will be re-used for the calculation of the constraints. Otherwise, if the statement $\boldsymbol{\mathcal{L}}_s \geq \kappa_3$ is true, then the information pairs used to calculate the $\boldsymbol{\mathcal{L}}_s$ will be re-used. If both statements are true, then the information pairs used for calculating $\boldsymbol{\mathcal{L}}_c$ will be re-used.
Once partial localizability is detected for any direction, the correspondence re-sampling process is employed, which aims to find the information pairs with the highest localizability contribution in the direction of the eigenvector. These pairs are then used to calculate a reliable pose estimate. This is achieved by solving a simplified least-squares minimization problem of the re-sampled pairs (cf. Equation~\ref{eq:solving_translation_resample} and~\ref{eq:solving_rotation_resample}), yielding the constraint values $\boldsymbol{t}_0, \boldsymbol{r}_0$. The overall re-sampling process is as follows:
\begin{enumerate}
    \item Acquire the degenerate eigenvector, $\boldsymbol{v}_j$ and the previously calculated localizability contribution vectors, $\mathcal{L}_s$ and $\mathcal{L}_c$. 
    \item Decide on the number of pairs to sample based on $\boldsymbol{\mathcal{L}}_c \geq \kappa_2$ and $\boldsymbol{\mathcal{L}}_s \geq \kappa_3$ conditions.
    \item Select the pairs based on their localizability contribution values. The localizability contribution value calculation is explained in depth in Section~\ref{task:Information_Analysis}.
    \item Based on the sub-space of $\boldsymbol{v}_j$, the simplified minimization problem in either Equation~\eqref{eq:solving_translation_resample} or~\eqref{eq:solving_rotation_resample} is solved.
\end{enumerate}
For the case of $\boldsymbol{v}_j \in \boldsymbol{V}_t$, the constraint value $\boldsymbol{t}_0$ can be computed by using the re-sampled pairs and solving the following minimization problem:
\begin{align}
\begin{gathered}
\displaystyle{{}^{re}_{}}{\boldsymbol{A}_t} = [{}^{re}_{}\boldsymbol{n}] \displaystyle[{}^{re}_{}\boldsymbol{n}]^\top, \quad \displaystyle{}^{re}_{}\boldsymbol{b}_t = [{}^{re}_{}\boldsymbol{n}][{}^{re}_{}\boldsymbol{n}]^\top
\displaystyle  \big({}^{re}_{}\boldsymbol{q}-{}^{re}_{}\boldsymbol{p}\big)\\
\displaystyle\mathop{\text{min}}_{\boldsymbol{t}_{0}\in \mathbb{R}^{3}} 
\displaystyle\Big|\Big|{}^{re}_{}\boldsymbol{A}_t
\displaystyle\boldsymbol{t}_{0}-{}^{re}_{}\boldsymbol{b}_t\Big|\Big|_2.
\label{eq:solving_translation_resample}
\end{gathered}
\end{align}
Here, the re-sampled information pairs are denoted as $\big\{{}^{re}_{}\boldsymbol{p},\{{}^{re}_{}\boldsymbol{n},{}^{re}_{}\boldsymbol{q}\} \big\} \in \mathcal{M}$. Similarly, for the case of $\boldsymbol{v}_j \in \boldsymbol{V}_r$ the constraint value $\boldsymbol{r}_0$ is calculated as follows:
\begin{align}
\begin{gathered}
\displaystyle{}^{re}_{}\boldsymbol{A}_r = \displaystyle[{}^{re}_{}\boldsymbol{p}\times{}^{re}_{}\boldsymbol{n}] \displaystyle[{}^{re}_{}\boldsymbol{p}\times{}^{re}_{}\boldsymbol{n}]^\top\\ {}^{re}_{}\boldsymbol{b}_r = \displaystyle[{}^{re}_{}\boldsymbol{p}\times{}^{re}_{}\boldsymbol{n}]\displaystyle[{}^{re}_{}\boldsymbol{n}]^\top
\displaystyle  \big({}^{re}_{}\boldsymbol{q}-{}^{re}_{}\boldsymbol{p}\big)\\
\displaystyle\mathop{\text{min}}_{\boldsymbol{r}_{0}\in \mathbb{R}^{3}} 
\displaystyle\Big|\Big|{}^{re}_{}\boldsymbol{A}_r
\displaystyle\boldsymbol{r}_{0}-{}^{re}_{}\boldsymbol{b}_r\Big|\Big|_2.
\label{eq:solving_rotation_resample}
\end{gathered}
\end{align}
The outputs $\boldsymbol{t}_0$ or $\boldsymbol{r}_0$ are the motion estimates based on the re-sampled correspondences in the direction of the degenerate eigenvector $\boldsymbol{v}_j$. As described in Section~\ref{section:detection}, partial localizability indicates available but sparse information along the degenerate directions, leveraged by the re-sampling process in the form of less-noisy and more reliable correspondences. 

However, as the data used in these minimization problems are deliberately selected to provide information in a specific direction, the Hessian matrices $\displaystyle{}^{re}_{}\boldsymbol{A}_t$ and $\displaystyle{}^{re}_{}\boldsymbol{A}_r $ might not be well-conditioned in some cases. 
To mitigate this risk, first, the Hessian matrices are factorized via LU decomposition with pivoting~\cite{lu}. Here the pivoting operation increases accuracy by interchanging rows to make the pivot element larger than any element below. 
Next, to reduce the possible adverse effects of ill-conditioning of the factorized problem, RIF preconditioning~\cite{RIF} is applied. An alternative to this would be the sampling of points in the well-constrained directions~\cite{gelfand2003} in addition to the degenerate directions, improving the conditioning of the Hessian. Yet, this re-sampling is not performed here due to its additional computational burden. 
Without these precautions, the calculated constraints from the re-sampling might not be accurate, and the residuals generated through projection onto the eigenvectors might negatively affect the convergence of the well-constrained directions.

\subsection{Constrained Optimization}
\label{task:contrained_optimization}
In the final step, the calculated constraint values $\boldsymbol{t}_0$ and $\boldsymbol{r}_0$ are incorporated into the least squares minimization problem. In order to be applicable, the individual constraints of Equation~\eqref{eq:constraint_form} must be extended and re-arranged to 6D: 
\begin{align}
\begin{split}
    \displaystyle\big[\boldsymbol{0}_{1\times3},\boldsymbol{v}_j\big]\cdot \boldsymbol{x}=\boldsymbol{v}_j\cdot\boldsymbol{t}_{0},\quad& \text{if } \displaystyle\boldsymbol{v}_j \in \boldsymbol{V}_t,\\
    \displaystyle\big[\boldsymbol{v}_j, \boldsymbol{0}_{1\times3}\big]\cdot \displaystyle\boldsymbol{x}=\boldsymbol{v}_j\cdot\boldsymbol{r}_{0},\quad& \text{if } \displaystyle\boldsymbol{v}_j \in \boldsymbol{V}_r.
\end{split}
 \label{eq:6d_constraint_form}
\end{align} 
Next, the constraints are brought to a matrix form of $\boldsymbol{C}\boldsymbol{x}=\boldsymbol{d}$. The number of constraints corresponds to the amount of \textit{none} and \textit{partial} categories along all eigenvectors, and is denoted as $m_t$ and $m_r$, with the total number of constraints $c=m_t+m_r \leq 6$. The final constraint matrix is
\begin{align}
\displaystyle\underbrace{\left[\begin{array}{ccc}
  \boldsymbol{0}_{m_r\times3}&    & \boldsymbol{v}_j \\
 \vdots &    &  \vdots \\
 \boldsymbol{v}_j &    &\boldsymbol{0}_{m_t\times3} \\
\end{array}\right]}_{\boldsymbol{C}_{(m_r + m_t) \times 6}}
\displaystyle\boldsymbol{x}=
    \underbrace{\left[\begin{array}{c}
    \boldsymbol{v}_j\cdot\boldsymbol{r}_0 \\
    \vdots\\
    \boldsymbol{v}_j\cdot\boldsymbol{t}_0 \\
    \end{array}\right]}_{\boldsymbol{d}_{(m_r + m_t) \times 1}},
\end{align}
where each row indicates an equality constraint. The whole optimization can then be re-expressed as
\begin{align}
\begin{split}
\mathop{\text{min}}_{\boldsymbol{x}\in \mathbb{R}^{6}}  \quad & \Big|\Big|\boldsymbol{A}^\prime
\boldsymbol{x}- \boldsymbol{b}^\prime\Big|\Big|_2,\\
\textrm{s.t.} \quad & \boldsymbol{C}\boldsymbol{x}-\boldsymbol{d}=0,
\end{split}
\label{eq:constrained_final_cost}
\end{align}
with $\boldsymbol{C}\in\mathbb{R}^{c\times6}$ and $\boldsymbol{d} \in \mathbb{R}^{c\times1}$.
Problem~\eqref{eq:constrained_final_cost} can be transformed into an unconstrained optimization problem by introducing Lagrangian multipliers~\cite{lagrangianMultipliers}, resulting in the final augmented linear least squares minimization problem
\begin{align}
\mathop{\text{min}}_{\boldsymbol{x}^\prime\in \mathbb{R}^{6}} 
\Big|\Big|\boldsymbol{A}^{\prime\prime}
\boldsymbol{x}^\prime- \boldsymbol{b}^{\prime\prime}\Big|\Big|_2,
\label{eq:augmented_langrange_multi}
\end{align}
with augmented optimization vector $\boldsymbol{x}^{\prime} = [\boldsymbol{{x}^{*}}^\top,\ \boldsymbol{{\lambda}^{*}}^\top]^\top$ with Lagrangian multipliers $\boldsymbol{\lambda} \in \mathbb{R}^{c\times1}$, an example of such an optimization problem is visualized in \mbox{Fig.~\ref{fig:optimizationModuleOverview}-B} for an optimization problem with a single equality constraint. 
The augmented matrices in Equation~\eqref{eq:augmented_langrange_multi} are defined as:
\begin{align*}
\underbrace{\left[\begin{array}{cc}
2\boldsymbol{A}^{\top} \boldsymbol{A} & \boldsymbol{C}^{\top} \\
\boldsymbol{C} & \boldsymbol{0}
\end{array}\right]}_{\boldsymbol{A^{\prime\prime}}}
\underbrace{\left[ \begin{array}{c}
\boldsymbol{x}^{*} \\
\boldsymbol{\lambda}^{*}
\end{array}\right]}_{\boldsymbol{x}^{\prime}}=
\underbrace{\left[\begin{array}{c}
2\boldsymbol{A}^{\top} \boldsymbol{b} \\
\boldsymbol{d}
\end{array}\right]}_{\boldsymbol{b}^{\prime\prime}}.
\end{align*}
This optimization problem can be solved via SVD, providing the optimal pose estimation ${\boldsymbol{x}^\prime}^*$ for the current ICP iteration.

It should be noted that in the case of a truly bad initial guess, the \mbox{Opt.-Module} will not be able to solve the point cloud registration problem reliably. Two factors contribute to this: i) inaccuracies in the correspondence search, affected by the bad initial guess. ii) In cases of complete degeneracy, the non-updated optimization directions will use the initial guess, which may result in an incorrect point cloud registration. While this is the case, this initial guess is utilized for all methods throughout this work and hence, affects all compared methods alike.



\section{Results} \label{section:results}
In this section, the experimental setup is discussed in Section~\ref{subsection:testingSetup}, followed by the performance evaluation of the proposed framework through Sections~\ref{subsection:simulation_dataset} to~\ref{subsection:opfikon_dataset}. The attached video\textsuperscript{\ref{foot:supplementary_material}} demonstrates the robot field deployment and summarizes the proposed framework. Finally, to validate the efficacy of individual sub-modules of the proposed solution, ablation studies are presented in Section~\ref{subsection:ablationStudy}.

\begin{figure}[t]
\centering
\includegraphics[width=1.0\linewidth]{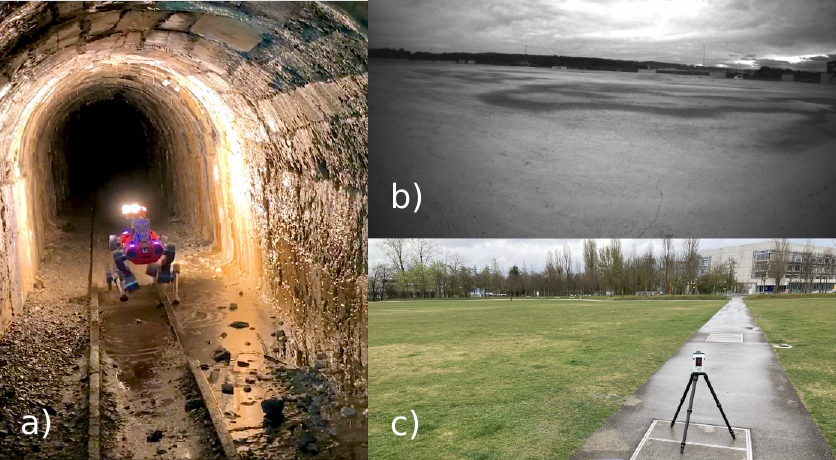}
\caption{Overview of the three real-world experiment sites: a) ANYmal in the \textit{Seem\"uhle Underground Mine} (\ref{sec:seemuehle}), b) an on-board image of the \textit{R\"umlang Construction Site} (\ref{sec:ruemlang}) showing the featureless planarity, and c) the ground truth collection at \textit{Opfikon City Park} (\ref{subsection:opfikon_dataset}) with the RTC360~\cite{RTC360} sensor.}
\label{fig:anymalInEnvironments}
\vspace{-1.0em}
\end{figure}

\subsection{Hardware \& Implementation Details} \label{subsection:testingSetup}
The proposed localizability-aware ICP framework is integrated into a modified C++ point cloud registration framework~\cite{pharosHiltiReport} developed by ANYbotics, which is based on the open-source registration library \textit{libpointmatcher}~\cite{libpointmatcher}, to demonstrate its suitability for challenging real-world applications.
An ANYmal-C~\cite{ANYmal} legged robot, shown in Fig.~\ref{fig:seemuhle_vlp16_result} and Fig.~\ref{fig:anymalInEnvironments}-(a), equipped with a Velodyne \mbox{VLP-16 LiDAR}, an inertial measurement unit (IMU), and joint encoders, was used in all field experiments. The initial transformation $\boldsymbol{T}_{\mathtt{M}\mathtt{L}, \text{init}}$ for point cloud registration and correspondence search is provided by ANYmal's leg odometry module~\cite{tsif}, which utilizes the IMU and joint encoder measurements. As discussed in~\cite{tsif}, the performance of this legged odometry module depends on the contact estimation performance, which is known to suffer on rough terrain. Finally, the point cloud motion compensation is performed at the driver level using the same odometry pose estimates in the transformation tree. All evaluations are performed on a laptop equipped with an \textit{Intel i7-9750H} CPU, equivalent to that available on the robot.

\subsection{Algorithmic Comparisons} \label{subsection:sota}
To facilitate a fair comparison with \mbox{X-ICP}, the current compared state-of-the-art methods (Zhang et al.~\cite{solRemap} and Hinduja et al.~\cite{hinduja2019degeneracy}) are re-implemented within the same ICP registration pipeline. The method of Zhang et al.~\cite{solRemap} requires an eigenvalue threshold for degeneracy detection, which is empirically set to $120$ to ensure good degeneracy detection for all experiments except for the Opfikon Park dataset in Section~\ref{subsection:opfikon_dataset}, where multiple eigenvalue threshold values are compared. The proposed framework utilizes the same localizability parameters $\kappa_{\{1,2,3\}}$ for all experiments.

\begin{figure}[t]
\includegraphics[width=\linewidth]{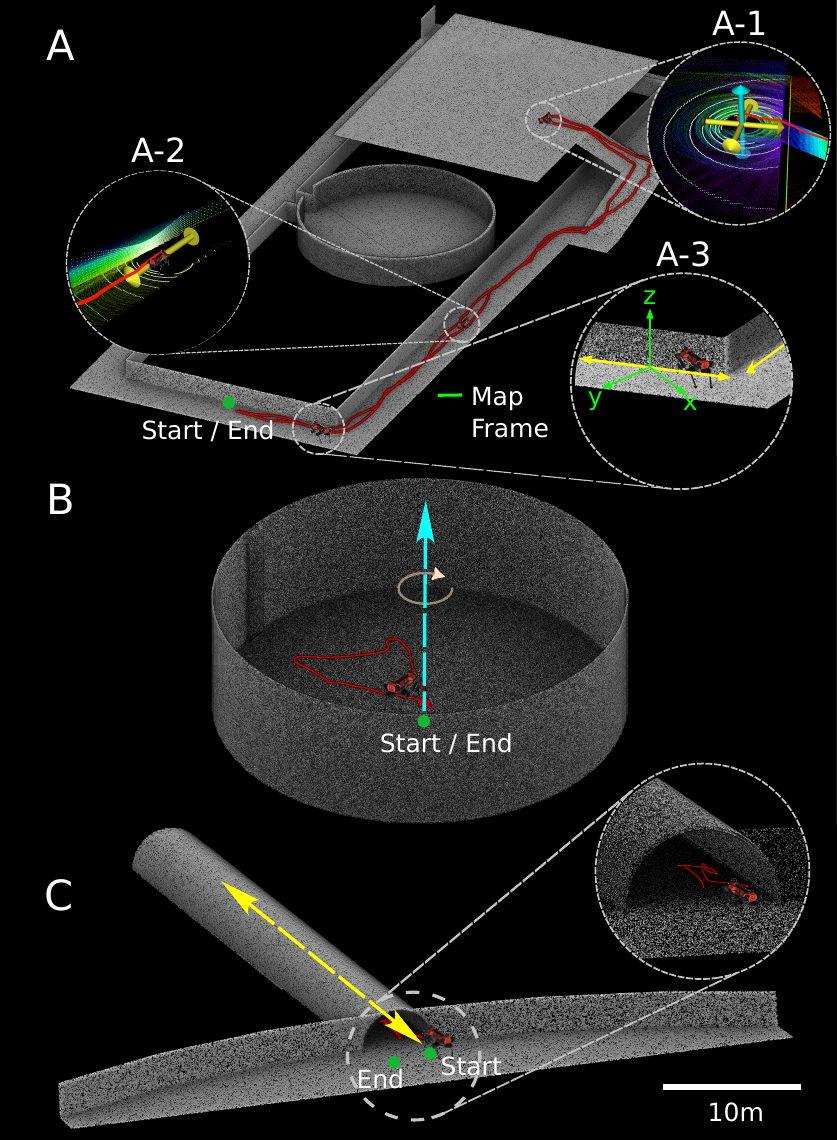}
\caption{An illustration of the three simulated environments of Section~\ref{subsection:simulation_dataset}. The degenerate directions are shown in yellow (translational) and blue (rotational) for all sub-figures, whereas the robot path is shown in red. The ground truth point clouds of the combined degeneracy (\textbf{A}), the rotational degeneracy (\textbf{B}), and the translational degeneracy (\textbf{C}) simulation environments are depicted above. Furthermore, \textbf{A-1} is a snapshot of \mbox{X-ICP}'s detection of 3-axes degeneracy, \textbf{A-2} is a snapshot of \mbox{X-ICP} being exposed to a single-axis translational degeneracy, while \textbf{A-3} shows the misalignment between the map frame and the degeneracy direction, highlighting the importance of direction-invariant localizability detection in eigenspace. 
}
\label{fig:simulation_dataset}
\vspace{-1.0em}
\end{figure}

\subsection{Simulation Study}
\label{subsection:simulation_dataset}
To evaluate the localizability detection performance of the proposed method, a set of tests are performed in simulation for better control over the quantities affecting the ICP registration, such as the quality of the point cloud registration prior. In addition, the designed simulation environments, shown in Fig.~\ref{fig:simulation_dataset}, are designed to feature smooth planar surfaces, self-similar corridors, a cylindrical room, and an open area - all are conditions known to induce ICP degeneracy.

\subsubsection{Translational Degeneracy}
\begin{figure}[t]
 \includegraphics[width=\linewidth]{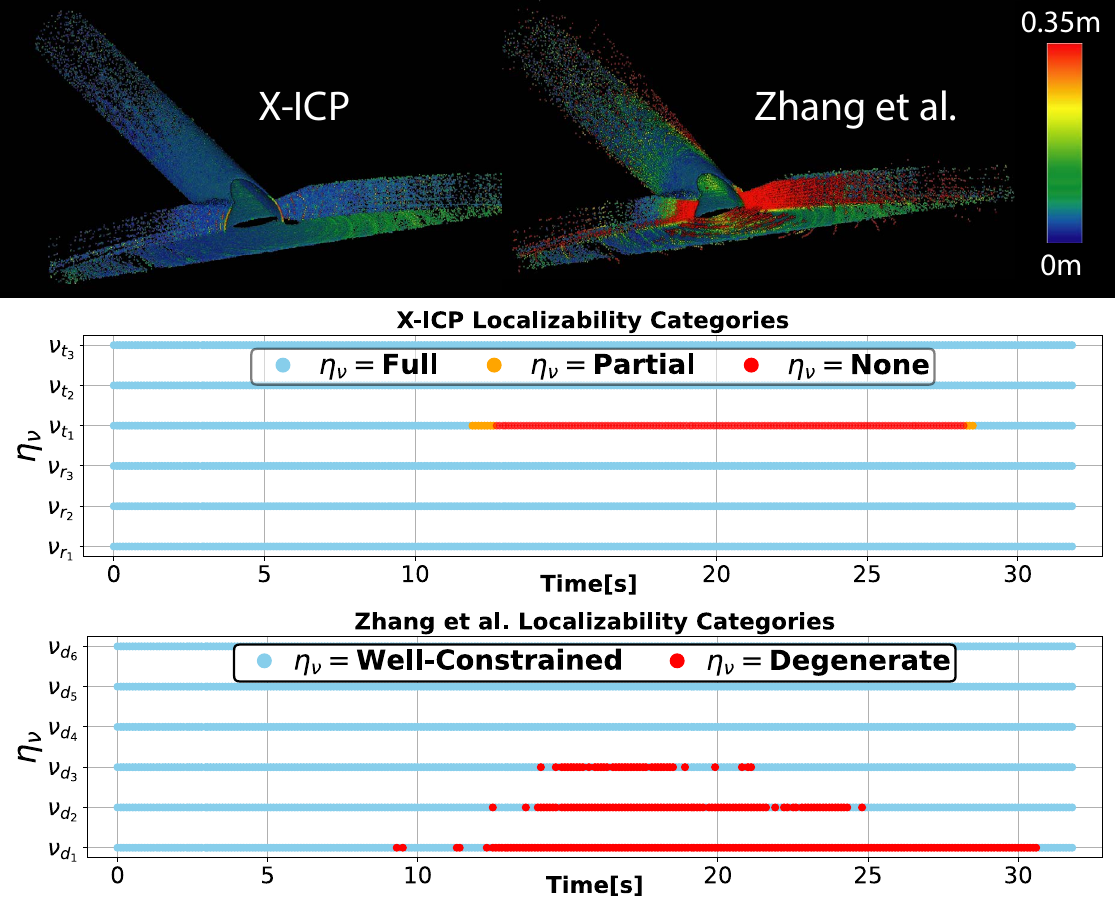}
\caption{\textbf{Top:} The point cloud maps produced using \mbox{X-ICP} and Zhang et al. methods for the translational degeneracy dataset. The color bar indicates the point-to-point distance error with respect to the ground truth map. \textbf{Bottom:} Plots showing the estimated localizability categories for both methods.}
\label{fig:simulationTranslation}
\vspace{-1.0em}
\end{figure}

In this study, pure translational degeneracy is investigated with a semi-circular tunnel environment with degeneracy along the longitudinal direction of the tunnel. This environment is shown in Fig.~\ref{fig:simulation_dataset}-C, with the robot path depicted in red. To penalize over-reliance on the legged odometry pose prior, noise is added to the initial guess before being given to the ICP algorithm for all methods. The noise is sampled from a velocity-dependent normal distributions $\mathcal{N}(\mu_t,\,{\sigma_t}^{2})$ and $\mathcal{N}(\mu_r,\,{\sigma_r}^{2})$. As an example, for velocities \SI{0.5}{\meter/\second} and \SI{0.2}{\radian/\second} the distribution variables would be $\mu_t=\SI{0}{\cm}, \ \sigma_t=\SI{0.0125}{\meter}$, $\mu_r=\SI{0}{\radian}, \ \sigma_r=\SI{0.005}{\radian}$. The mapping results of this study, including an error map and the predicted localizability categories, are shown in Fig.~\ref{fig:simulationTranslation}. In this example, \mbox{X-ICP} is able to identify the degeneracy correctly along one direction, despite given the noisy registration prior. The localizability detection of Zhang et al.~\cite{solRemap} detects degeneracy along three directions, which is incorrect. Furthermore, as seen from the error map, the generated map using Zhang et al.'s~\cite{solRemap} method shows higher error and drift along the principal direction of the tunnel, demonstrating the advantages of \mbox{X-ICP}, which can successfully identify and mitigate translational degeneracy.

\subsubsection{Rotational Degeneracy}
\begin{figure}[t]
\includegraphics[width=\linewidth]{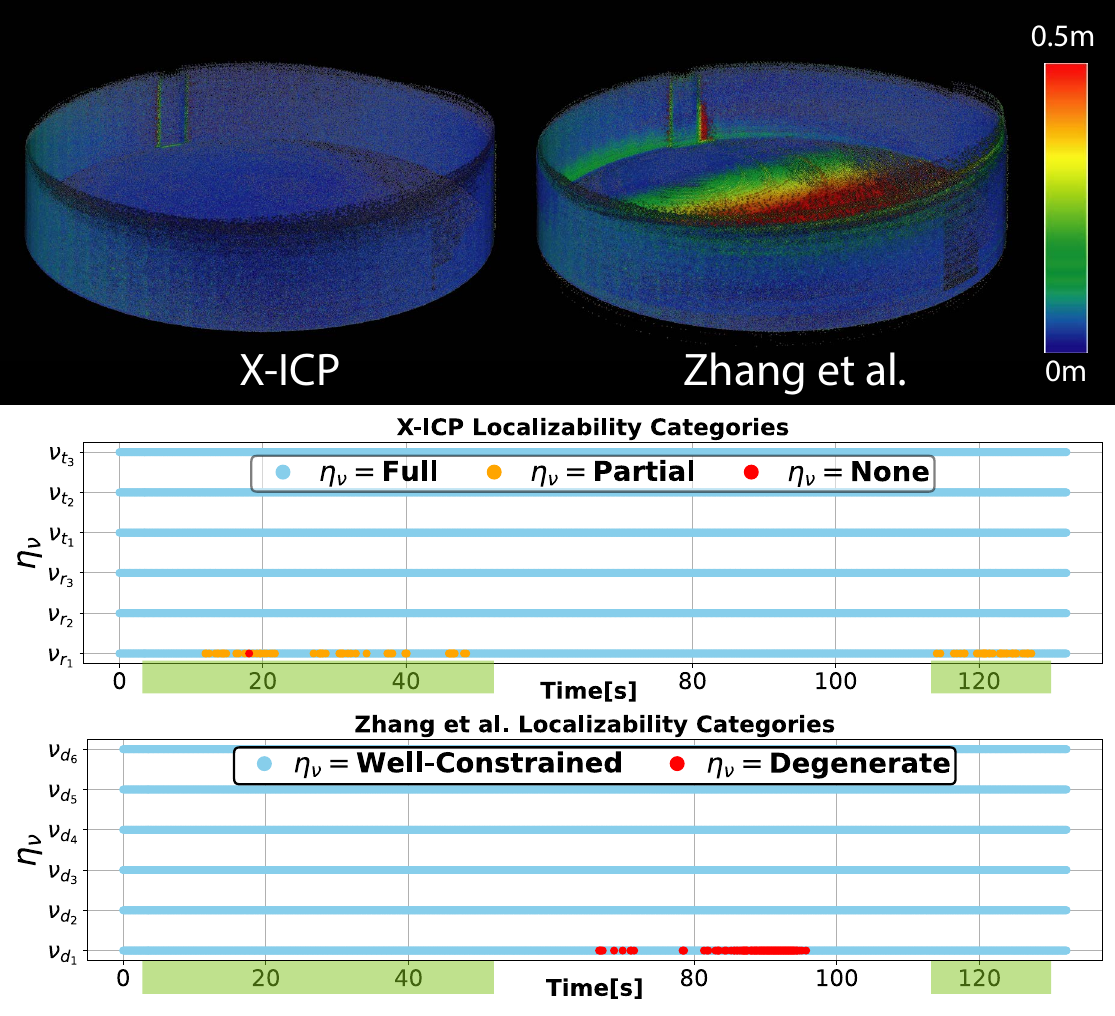}
\caption{ \textbf{Top:} The point cloud maps produced using \mbox{X-ICP} and method from Zhang et al.~\cite{solRemap} for the rotational degeneracy dataset. The color bar indicates the point-to-point distance error. \textbf{Bottom:} Plots showing the estimated localizability categories for both methods. The time interval of the robot being close to the center of the cylinder is highlighted as a shaded green area.}
\label{fig:simulationRotation}
\vspace{-1.0em}
\end{figure}

The rotational degeneracy is simulated in a cylindrical environment (cf. Fig.~\ref{fig:simulation_dataset}-B) with known degeneracy for rotation axes perpendicular to the cylinder's ground surface. Similar to the translational degeneracy test, a motion-based odometry noise is added to the initial prior provided to the ICP algorithm. The results of this study are shown in Fig.~\ref{fig:simulationRotation}. It can be seen that \mbox{X-ICP} identifies the rotational degeneracy correctly, and the resulting point cloud map contains small errors. \mbox{X-ICP}'s degeneracy detection nicely overlaps with the time intervals of the robot walking near the center of the environment, indicated as green overlays in Fig.~\ref{fig:simulationRotation}. In contrast, the localizability detection of Zhang et al.~\cite{solRemap} indicates degeneracy when the robot is near the circular wall, which is incorrect. This underlines the sensitivity of the detection algorithm w.r.t. the observed number of points.
Furthermore, as seen from the map error comparison, the map produced using Zhang et al.'s~\cite{solRemap} method shows higher error when compared to the map produced by \mbox{X-ICP}, demonstrating its efficacy for operation in rotation degenerate environments.

\subsubsection{Combined Degeneracy}
\begin{figure}[t]
\centering
\includegraphics[width=1.0\linewidth]{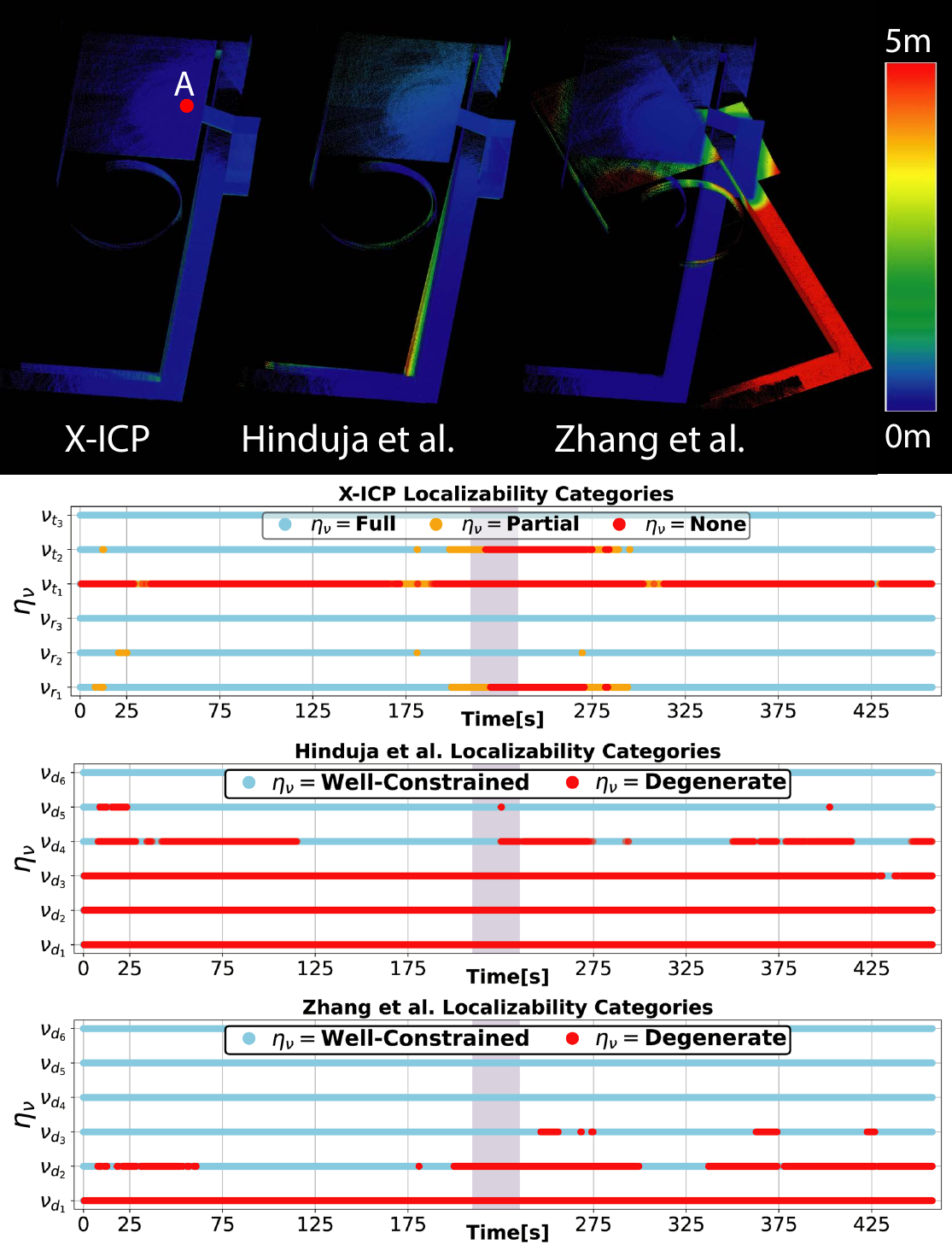}
\caption{\textbf{Top:} The generated point cloud maps of the three methods for the combined simulation environment. The color bar indicates the point-to-point distance error of the produced maps. \textbf{Bottom:} The estimated localizability categories of \mbox{X-ICP} and the state-of-the-art methods are shown. The region corresponding to point \textbf{A} in the top figure is highlighted in the plot. }
\label{fig:simulation_results}
\vspace{-1.0em}
\end{figure}

In this experiment shown in Fig.~\ref{fig:simulation_dataset}-A, the robot starts in a corridor-like area, navigates to an open space (Fig.~\ref{fig:simulation_dataset}-A-1), and returns to the starting position, traversing a total distance of \SI{246}{\meter} at a nominal velocity of \SI{0.5}{\meter/\second}. In the corridor sections (shown in Fig.~\ref{fig:simulation_dataset}-A-2 and A-3), the ICP registration is expected to degenerate in one axis along the corridor. In contrast, in the open section (Fig.~\ref{fig:simulation_dataset}-A-1), the degeneracy is expected to occur along two translational directions parallel to the ground plane and in one rotational direction perpendicular to the ground plane. The results shown in Fig.~\ref{fig:simulation_results} validate the correct degeneracy detection performance of the proposed work; non-localizability is detected along one axis throughout, corresponding to the corridor sections. Furthermore, non-localizability along two additional axes is only detected for the 200-300\si{\second} interval, corresponding to the open section of the environment. The localizability detection of \mbox{X-ICP} is compared against the state-of-the-art methods~\cite{solRemap,hinduja2019degeneracy} in the bottom rows of Fig.~\ref{fig:simulation_results}, and a comparison of the three maps is presented in the top of Fig.~\ref{fig:simulation_results}.
As discussed in the introduction of the simulation study, to eliminate the effect of an imperfect initial guess, in this simulation experiment, a perfect prior is fed to all three methods. 
The method of Hinduja et al.~\cite{hinduja2019degeneracy} remains overly pessimistic and fully relies on the ICP pose prior for the registration, resulting in less drift than Zhang et al.~\cite{solRemap}, which in contrast remains overly optimistic in the detection of rotation degeneracy, leading to a broken map. The section where Zhang et al.~\cite{solRemap} extensively drifts is indicated with point \textbf{A} in Fig.~\ref{fig:simulation_results} and the corresponding time frame is highlighted with the shaded area on the localizability estimation plots.
The proposed method, which also utilizes the partial localizability information along degenerate directions, shows minimal map error w.r.t. the ground-truth map.

\begin{figure}[t]
\includegraphics[width=\linewidth]{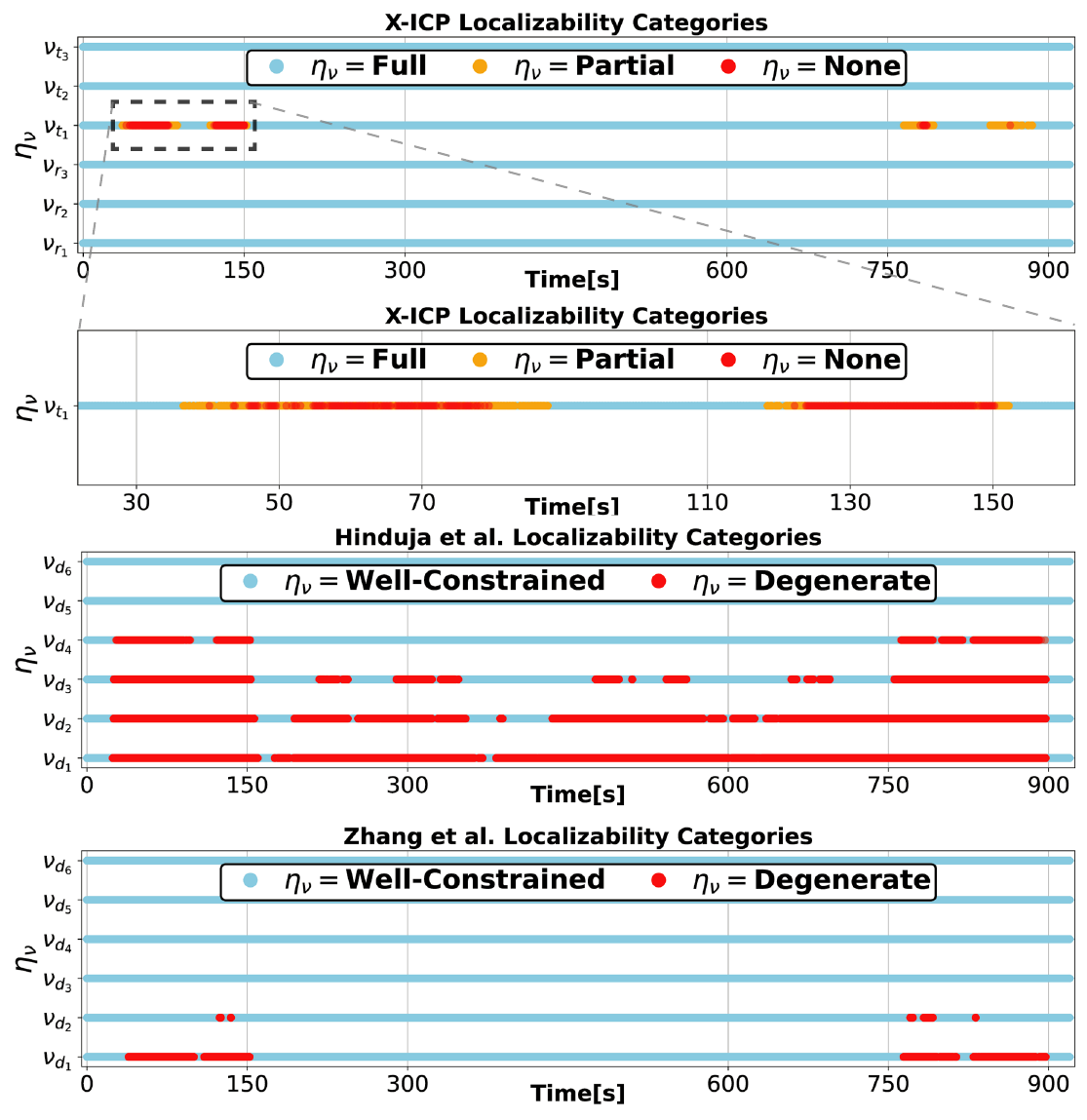}
\caption{Localizability predictions of different methods for the Seem\"uhle \mbox{VLP-16} experiment. The localizability categories by the proposed \mbox{X-ICP} (\textbf{top}), the method from Hinduja et al.~\cite{hinduja2019degeneracy} (\textbf{middle}), and of Zhang et al.~\cite{solRemap} (\textbf{bottom}) are shown.}
\label{fig:seemuhle_localizability_categories}
\vspace{-1.0em}
\end{figure}

\subsection{Seem\"uhle Underground Mine}
\label{sec:seemuehle}
Next, the proposed method is evaluated and studied in three real-world environments. In the first experiment, the ANYmal robot traversed an abandoned underground mine in Switzerland called Seem\"uhle. This environment constitutes several challenges: First, the ground of the mine is uneven due to sharp rocks and rails, leading to constant foot slippage, deteriorating the ground contact estimation and, thus, the odometry prior~\cite{tsif} for point cloud registration. Second, the environment contains a long tunnel segment with smooth arched walls (cf. Fig.~\ref{fig:anymalInEnvironments}-a and Fig.~\ref{fig:seemuhle_vlp16_result}-A, B, C) that do not provide reliable constraints along the principal direction of the tunnel. The environment was scanned using a Leica \mbox{RTC 360} scanner (cf. Fig.~\ref{fig:seemuhle_vlp16_result}), which is used to generate ground truth map and robot trajectory, the latter using the method in~\cite{newerCollegeDataset}. The same experiment was performed twice using the same ANYmal robot; once with the Velodyne \mbox{VLP-16} LiDAR and once with the Ouster \mbox{OS0-128} LiDAR, allowing for a comparative analysis of the effect of different point cloud density configurations on the proposed method. The robot traversed a total distance of \SI{521.8}{\meter} during this experiment.

\subsubsection{Velodyne VLP-16 LiDAR}\label{vlp16seemuhlesubsub}
The error heat maps of the three methods for the experiment conducted using the Velodyne \mbox{VLP-16} LiDAR sensor are shown in the bottom row of Fig.~\ref{fig:seemuhle_vlp16_result}. 
Both of the compared state-of-the-art approaches suffer from the impact of environmental degeneracy, prohibiting a drift-free traversal. 
The method of Zhang et al.~\cite{solRemap} performs comparably to the proposed solution until entering the tunnel on the way back to the starting point. In this second part, the degeneracy detection works incorrectly, resulting in \mbox{LiDAR} slip. The localizability categories estimated by Zhang et al.~\cite{solRemap} are shown at the bottom of Fig.~\ref{fig:seemuhle_localizability_categories}, indicating the incorrect and pessimistic localizability estimation.
The solution of Hinduja et al.~\cite{hinduja2019degeneracy} shows a larger registration error throughout the experiment, as depicted in the error heat map, due to a pessimistic localizability estimation as shown in the middle figure of Fig.~\ref{fig:seemuhle_localizability_categories}. 
This causes over-relying on the noisy odometry prior instead of registration, thus creating an incorrect final point cloud map.

In contrast, \mbox{X-ICP} performs reliable and consistent registration throughout the tunnel segment, as demonstrated by the low error of the produced point cloud map in Fig.~\ref{fig:seemuhle_vlp16_result}. The corresponding localizability estimation is shown in the top plot of Fig.~\ref{fig:seemuhle_localizability_categories}, underlining the ability of the \mbox{X-ICP} \mbox{Loc.-Module} to capture the tunnel section at the beginning and end of the experiment. Here, the advantage of detecting localizability in eigenspace becomes clear; although the robot moves and rotates within the environment, the degeneracy only affects a single direction of the optimization, in this case, $\boldsymbol{v}_{t_1}$. Since this vector is defined in the eigenspace, it does not need to align with any of the Cartesian axes in the ICP or optimization frame. A close-up of the localizability detection in the tunnel is provided again in the top row of Fig.~\ref{fig:seemuhle_localizability_categories}. The localizability categorization in certain parts of the tunnels shows the ability to distinguish the subtle differences at the bending of the tunnel point (B in Fig.~\ref{fig:seemuhle_vlp16_result}), allowing the optimization to utilize the given information for better registration. 

The absolute pose error (APE), relative pose error (RPE)~\cite{ate_rpe}, and end position errors are calculated using the EVO evaluation package\footnote{\url{https://github.com/MichaelGrupp/evo}\label{foot:evo}}. APE is a measure of the global accuracy of the estimated robot trajectory.
Results for two different trajectory alignment methods are reported in Table~\ref{table:APE_results} for the cases of either \textit{i)} aligning the first \SI{15}{\meter} ($\approx$200 poses) of the trajectory, or \textit{ii)} aligning the first pose of the ground truth and the estimated poses.
\begin{table}[ht]
\caption{APE Error for \mbox{VLP-16} Seem\"uhle experiment  (best in \textbf{bold}).}
\resizebox{\columnwidth}{!}{
\begin{tabular}{lccccc}
\toprule[1pt]
\multirow{2}[3]{*}{} & \multicolumn{2}{c}{\makecell{First \SI{15}{\meter} Alignment}} & \multicolumn{3}{c}{Origin Alignment} \\
\cmidrule(lr){2-3}  \cmidrule(lr){4-6}
& \makecell{Translation \\ $\mu(\sigma)[m]$ } & \makecell{Rotation \\ $\mu(\sigma)[\deg]$} & \makecell{Translation \\ $\mu(\sigma)[m]$ } & \makecell{Rotation \\ $\mu(\sigma)[\deg]$} & \makecell{Last Position \\ Error[$m$]}\\
\toprule[1pt]
\makecell{X-ICP (Proposed)} &\textbf{2.05}(\textbf{1.23}) & \textbf{2.55}(\textbf{0.76}) & \textbf{2.45}(\textbf{1.35}) & \textbf{2.50}(\textbf{1.03}) & \textbf{0.27} \\[0.5em]
\makecell{\textit{Zhang et al.}~\cite{solRemap}} & 3.36(1.74) & 4.06(1.37) & 3.73(1.80) & 4.11(1.52) & 6.37 \\[0.5em]
\makecell{\textit{Hinduja et al.}~\cite{hinduja2019degeneracy}}&   5.79(5.26)  &  7.67(4.72) & 8.16(4.83) &8.03(4.73) & 24.17 \\
\toprule[1pt]
\end{tabular}
}
\label{table:APE_results}
\vspace{-1.0em}
\end{table}
In addition to APE, the end position error is also calculated as the difference between the last estimated robot position and the ground truth position. Both the APE metric and the end translation errors in Table~\ref{table:APE_results} indicate that the compared state-of-the-art methods globally drift, whereas the proposed solution provides an accurate localization solution for real-world applications in challenging environments. The RPE error is measured in Table~\ref{table:RPE_results} to analyze local pose estimation accuracy and relative pose drift. The results demonstrate that the proposed method performs better than the state-of-the-art methods in local consistency.
\begin{table}[ht]\centering
\vspace{-1.0em}
\caption{RPE per \SI{10}{\meter} traversed distance for Seem\"uhle Mine experiment with \mbox{VLP-16} (best in \textbf{bold}).}
\begin{tabular}{lcc}
\toprule[1pt]
& \makecell{Translation \\ $\mu(\sigma)[m]$ } & \makecell{Rotation \\ $\mu(\sigma)[\deg]$}\\
\toprule[1pt]
\makecell{X-ICP (Proposed)} &\textbf{0.17}(\textbf{0.12}) &\textbf{0.86}(\textbf{0.42}) \\[0.5em]
\makecell{\textit{Zhang et al.}~\cite{solRemap}} & 0.20(0.14) & 0.93(0.51)\\[0.5em]
\makecell{\textit{Hinduja et al.}~\cite{hinduja2019degeneracy}}& 0.26(0.14)  &1.28(0.74)\\
\toprule[1pt]
\end{tabular}
\label{table:RPE_results}
\vspace{-1.0em}
\end{table}

\begin{figure}[!t]
\includegraphics[width=\linewidth]{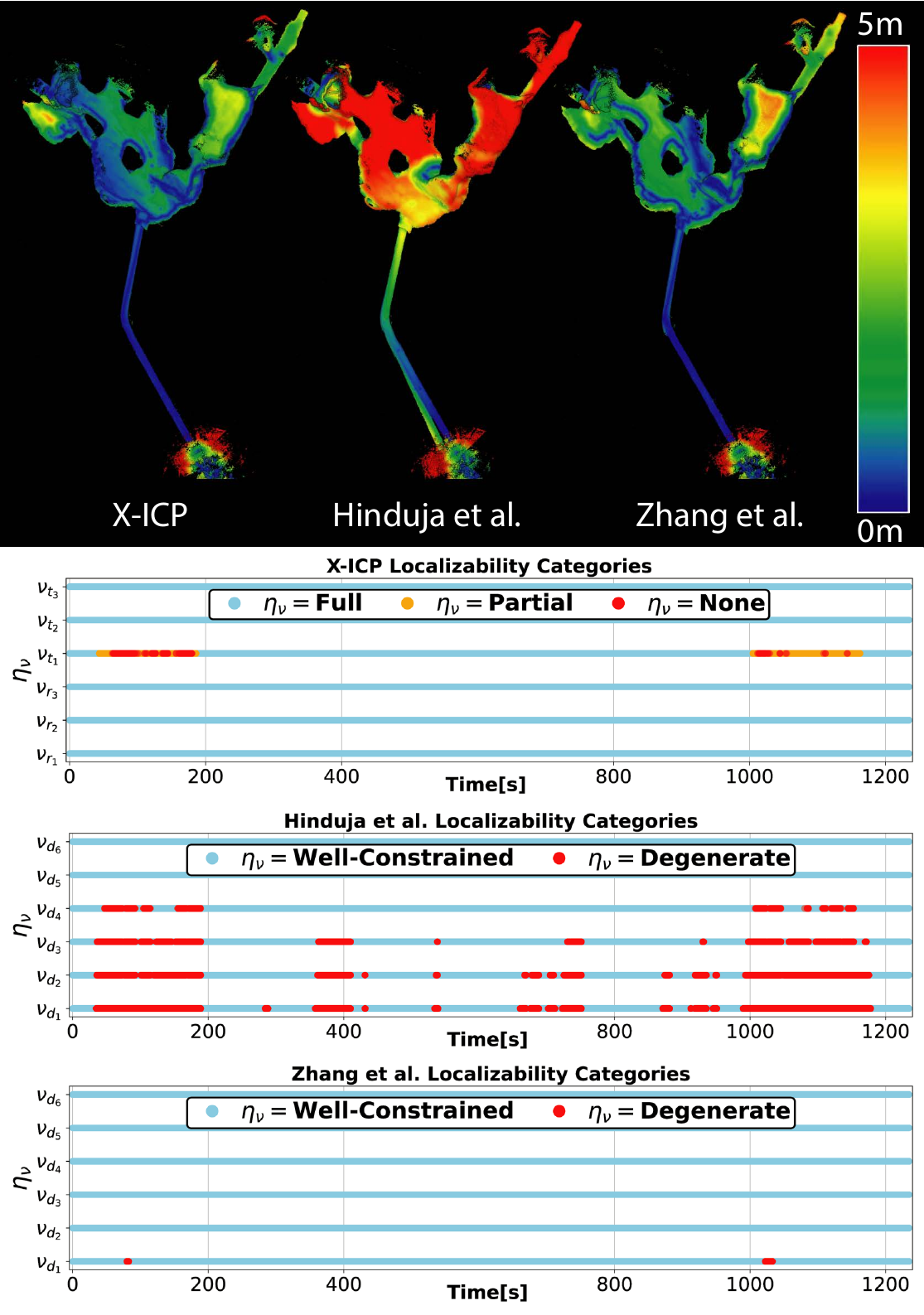}
\caption{\textbf{Top Row:} Point cloud maps of the three approaches for the  Ouster \mbox{OS0-128} Seem\"uhle experiment. The color bar indicates the point-to-point distance error. \textbf{Bottom Row:} The estimated localizability categories of \mbox{X-ICP} and the state-of-the-art methods are shown.}
\label{fig:seemuhle_os128_result}
\vspace{-1.0em}
\end{figure}

\subsubsection{Ouster OS0-128}
To investigate the robustness and applicability of the proposed approach to different sensor setups, the experiment is repeated using an Ouster \mbox{OS0-128} LiDAR with a much higher point density and larger FoV. To accommodate the increase in sensor noise, the filtering threshold $\kappa_f$ is reduced to $60^\circ$. As seen in the top row of Fig.~\ref{fig:seemuhle_os128_result}, all three approaches benefit from this higher density in data and perform better than their performance with the \mbox{VLP-16} LiDAR data. Despite the improvement, the solution from Hinduja et al.~\cite{hinduja2019degeneracy} still performs sub-optimally and generates a map with visible drift. On the other hand, Zhang et al.~\cite{solRemap} can complete the trajectory without substantial drift. Moreover, the \mbox{X-ICP} solution shows less map point-to-point error in the inner parts of the cave. In addition, the localizability categories are shown at the bottom row of Fig.~\ref{fig:seemuhle_os128_result}. Similar to the experiment with \mbox{VLP-16}, the degeneracy in the tunnel section of the environment is well-captured by \mbox{X-ICP} while the solution from Hinduja et al.~\cite{hinduja2019degeneracy} estimates localizability pessimistically. Interestingly, the method of Zhang et al.~\cite{solRemap} fails to detect degeneracy in the tunnel section compared to the experiment with \mbox{VLP-16}. Previously, Nubert et al.~\cite{nubert2022learning} investigated and showed that the higher point density from Ouster \mbox{OS0-128} LiDAR results in bigger eigenvalues; hence, the localizability threshold of Zhang et al.~\cite{solRemap} requires heuristic tuning.

\begin{figure}[t]
\includegraphics[ width=\linewidth]{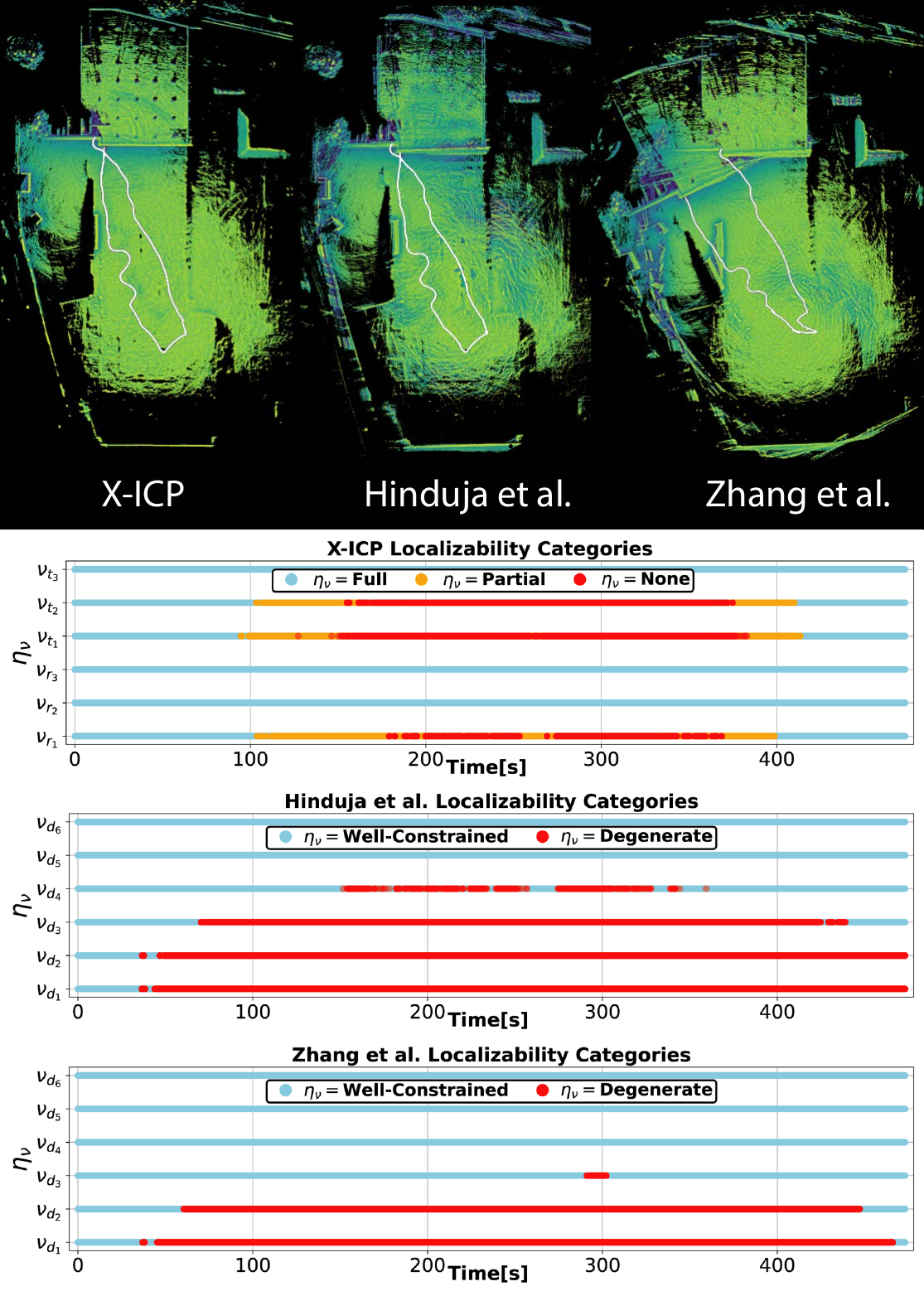}
\caption{\textbf{Top Row:} Point cloud maps of \mbox{X-ICP} and the two state-of-the-art approaches for the R\"umlang experiment. \textbf{Bottom:} The estimated localizability categories by \mbox{X-ICP} and the state-of-the-art methods are compared.}
\label{fig:rumlang_result}
\vspace{-1.0em}
\end{figure}
\subsection{R\"umlang Construction Site}
\label{sec:ruemlang}
In the next experiment, the robot navigated in a large-scale construction site in R\"umlang, Switzerland, posing an under-constrained scenario for translation along the ground plane and rotation perpendicular to it. A picture taken in this open environment is shown in Fig.~\ref{fig:anymalInEnvironments}-b), demonstrating its structureless planarity. During the \SI{153}{\meter} long test, the robot started next to a garage-like structure, traversing to the open area where it performed a couple of in-spot rotations before returning back. The in-spot rotations challenge the yaw estimation performance of the registration methods. 

The experimental results are presented in Fig.~\ref{fig:rumlang_result}, with the top row showing point cloud maps generated using the three different methods and the bottom row showing the associated localizability category estimation of different methods.
It can be seen that relying on the method by Zhang et al.~\cite{solRemap} leads to poor performance due to its reliance on the eigenvalues of the optimization Hessian matrix. In particular, the detection of rotation and translation degeneracy with a single threshold is difficult, as the scale of eigenvalues between rotation and translation sub-spaces differ significantly. 
The method by Hinduja et al.~\cite{hinduja2019degeneracy} performs better in this highly degenerate environment due to its pessimistic localizability detection for pose estimation, it relies more on the legged odometry prior, which performs well on the flat concrete ground. Nevertheless, upon closer observation, map distortion in the form of blurry rigid structures can be noted due to errors accumulated from the drift in the pose prior.
In contrast, the proposed method produces a consistent map of the environment preserving fine details, with its localizability detection performance shown in the second row of Fig.~\ref{fig:rumlang_result}. 

\begin{figure}[t]
\includegraphics[trim=0 175 0 0, clip, width=\linewidth]{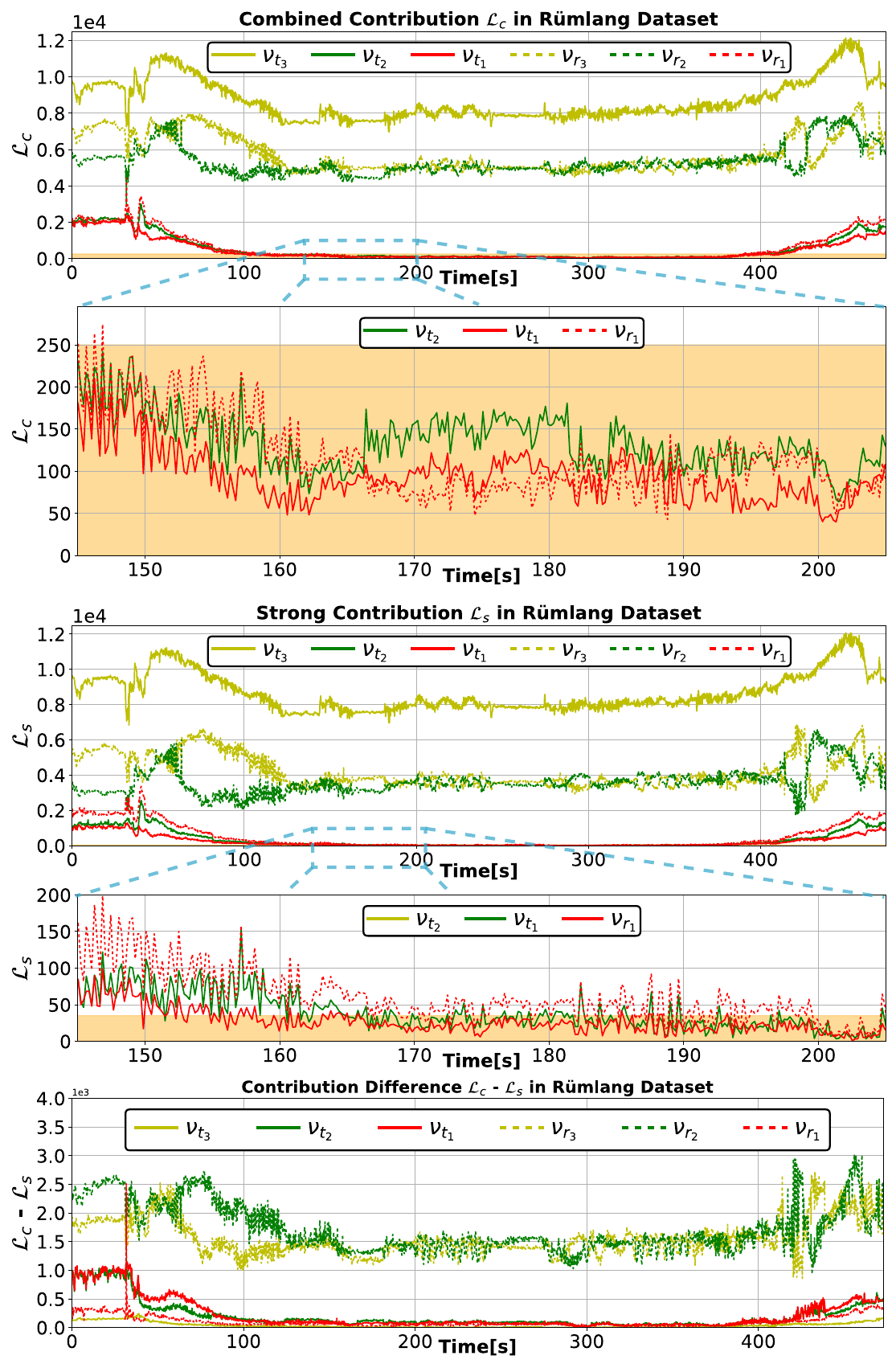}
\caption{\textbf{Top row:} Evolution of $\mathcal{L}_c$ throughout the R\"umlang experiment. The shaded orange region (zoomed-in \textbf{middle row}) represents the active region of the localizability awareness. \textbf{Bottom row:} The difference between $\mathcal{L}_c$ and $\mathcal{L}_s$ is provided to illustrate the filtered information pairs.}
\label{fig:rumlang_contributions}
\end{figure}

Finally, \mbox{X-ICP} not only correctly detects non-localizability along the three degenerate directions in the open area, but it also captures the smooth transition between partial and non-localizability. The smooth evolvement of the localizability categories can be better understood by observing the contribution changes provided in Fig.~\ref{fig:rumlang_contributions}. The three degenerate directions are easily identifiable with their comparably low combined contribution value. In addition, inspecting the strong contribution plot reveals direction $\boldsymbol{v}_{r_1}$ to be mostly partially localizable. This is inferred through the strong contribution peaks that exceed $\kappa_3 = 35$.


\subsection{Opfikon City Park}
\label{subsection:opfikon_dataset}
\begin{figure}[hb]
\centering
\includegraphics[width=\linewidth]{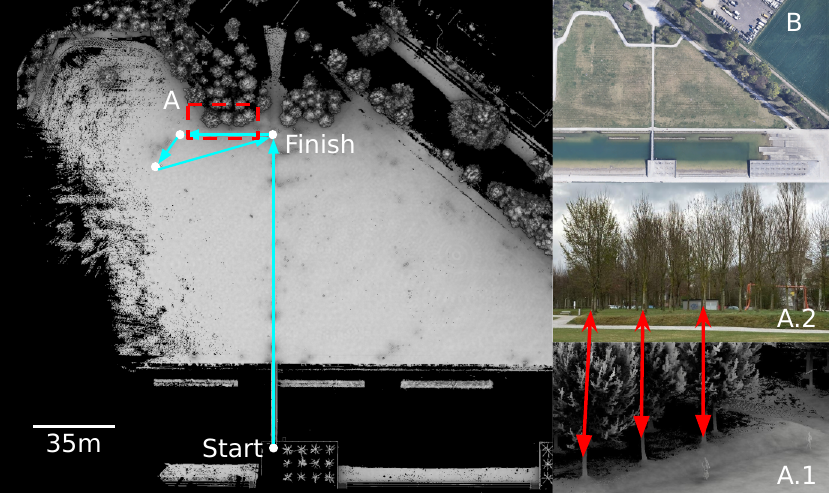}
\caption{Ground truth point cloud map of the Opfikon City Park. \textbf{A:} The highlighted trees are later investigated in Figure~\ref{fig:glattpark_results} as a demonstration of the effectiveness of \mbox{X-ICP}. \textbf{B:} Satellite view of the Opfikon City Park.}
\label{fig:glattpark_gt}
\vspace{-1.0em}
\end{figure}

To evaluate the efficacy of the proposed method in natural environments, an experiment was conducted in an outdoor park at Opfikon, Switzerland.
In this experiment, the robot in total traversed \SI{235}{\meter} over soft terrain and a suspension bridge, both adversely affecting the quality of the legged odometry prior. In addition, the vegetation renders reliable surface-normal extraction difficult. Beyond that, towards the end of the experiment, the robot enters an open unstructured area near the center of the park, posing a LiDAR degenerate situation. The ground truth map of the environment, shown in Fig.~\ref{fig:glattpark_gt}, is collected using a Leica \mbox{RTC 360} scanner (\mbox{Fig.~\ref{fig:anymalInEnvironments}-c}).  

\begin{figure}[hb]
\includegraphics[width=\linewidth]{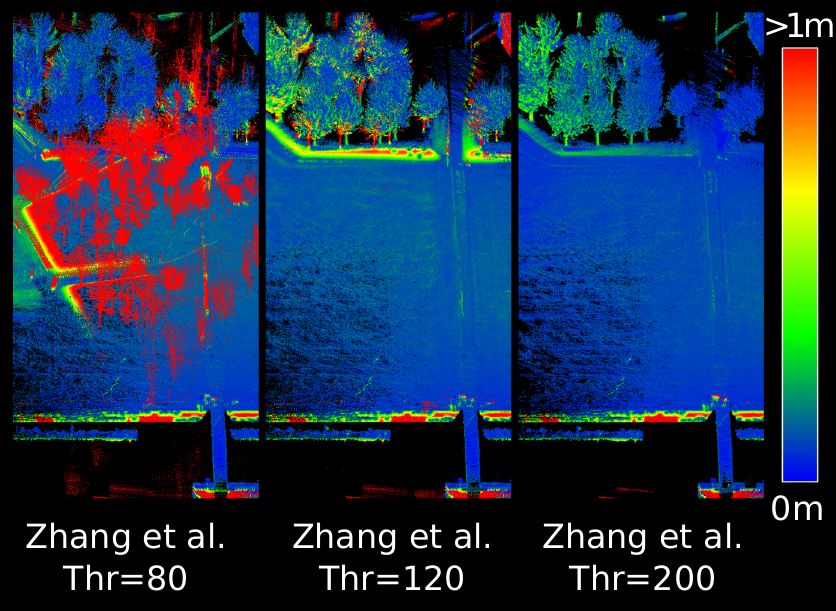}
\caption{Point cloud maps produced using the approach by Zhang et al.~\cite{solRemap} for different eigenvalue thresholds for the Opfikon experiment. The color bar indicates the point-to-point distance error w.r.t. the ground truth map.}
\label{fig:glattparkEigenValues}
\vspace{-1.0em}
\end{figure}

Unlike the previous tests, the degeneracy detection threshold of Zhang et al.~\cite{solRemap} is adapted to highlight the importance of heuristic tuning, as demonstrated in Fig.~\ref{fig:glattparkEigenValues}. A sub-optimal threshold degrades the quality of the map, and hence, in favor of~\cite{solRemap}, the threshold is tuned to be $200$ (instead of $120$).

\begin{figure}[t]
\includegraphics[width=\linewidth]{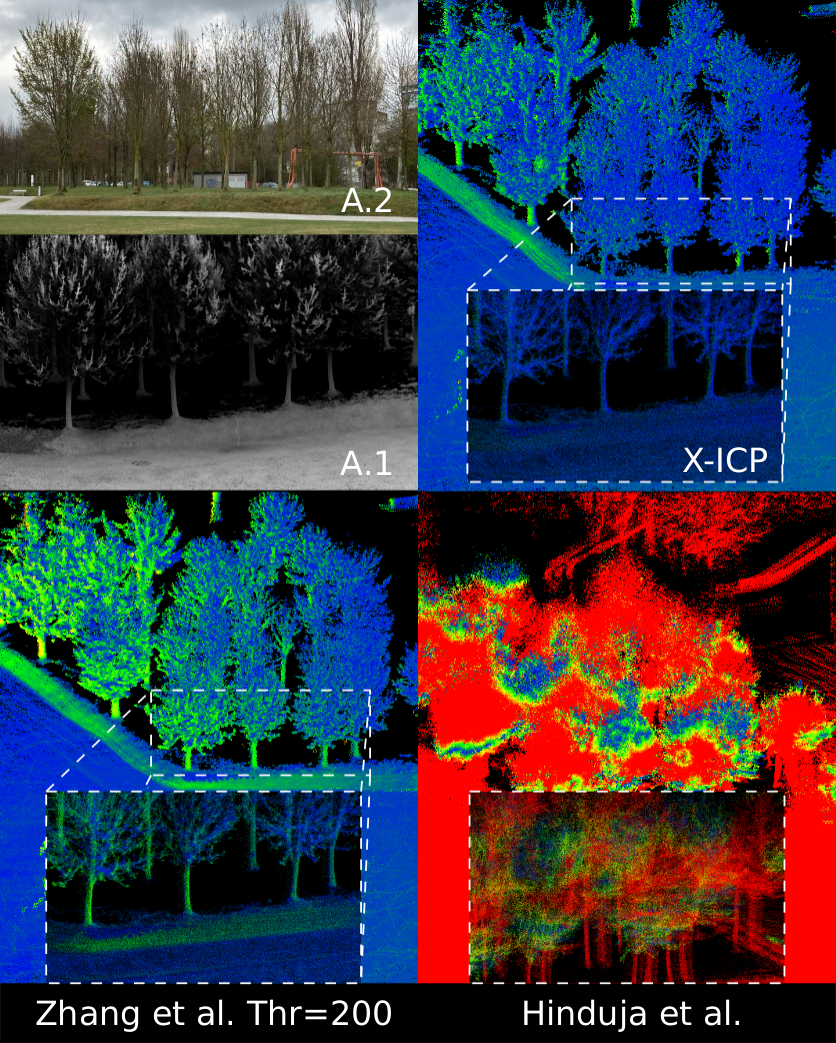}
\caption{Zoomed-in point cloud maps of the three investigated approaches for the Opfikon experiment (cf. Fig.~\ref{fig:glattpark_gt}). 
The error color scale is identical to Fig~\ref{fig:glattparkEigenValues}. \textbf{A.1} and \textbf{A.2} depict ground truth and real images for the tree region.}
\label{fig:glattpark_results}
\vspace{-1.0em}
\end{figure}

A map error evaluation of the three methods is presented in Fig.~\ref{fig:glattpark_results}. The magnified parts show the most accurate and fine-detailed map to be produced by the proposed method. Hinduja et al.~\cite{hinduja2019degeneracy} remains over-pessimistic and introduces significant drift in the presence of noisy odometry prior. Moreover, despite the tuned degeneracy detection threshold for Zhang et al.~\cite{solRemap}, the effect of LiDAR degeneracy is clearly visible in the tree section compared to the proposed method. This result concludes the evaluation of the proposed method across natural and man-made environments and emphasizes the advantage of the proposed (heuristic-free) localizability-aware registration.

\subsection{Ablation Studies} \label{subsection:ablationStudy}
The results discussed in the previous section validate the advantage of employing \mbox{X-ICP} in different LiDAR-challenging environments. 
To better understand the improvements, the effect of different components of the proposed method is investigated in an ablation study. In particular, a simplified version of \mbox{X-ICP}, referred to as \mbox{\textbf{Xs-ICP}}, is compared in view of fine-grained localizability categorization and the impact of performing the localizability analysis in every iteration of the ICP algorithm. In \mbox{Xs-ICP}, the partial-localizability category is discarded to reduce complexity, and the localizability parameters are set as $\kappa_1 \geq \kappa_2 = \kappa_3$, with values $\kappa_1=250$ and $ \kappa_2 = 180$. Without employing partial localizability, \mbox{Xs-ICP} performs localizability detection similar to the literature~\cite{solRemap,hinduja2019degeneracy} in a binary fashion.
The decision tree of \mbox{Xs-ICP} is also simplified, with comparisons of only $\boldsymbol{\mathcal{L}}_c \geq \kappa_1 \; \text{and} \; \boldsymbol{\mathcal{L}}_s \geq \kappa_3$. If either of these statements holds, the ICP optimization problem is well-constrained in the direction of the eigenvector. Finally, \mbox{Xs-ICP} performs the categorization step only in the first iteration of the ICP algorithm while in \mbox{X-ICP}, the categorization is performed at every iteration of the ICP algorithm.
\begin{figure}[t]
\includegraphics[width=\linewidth]{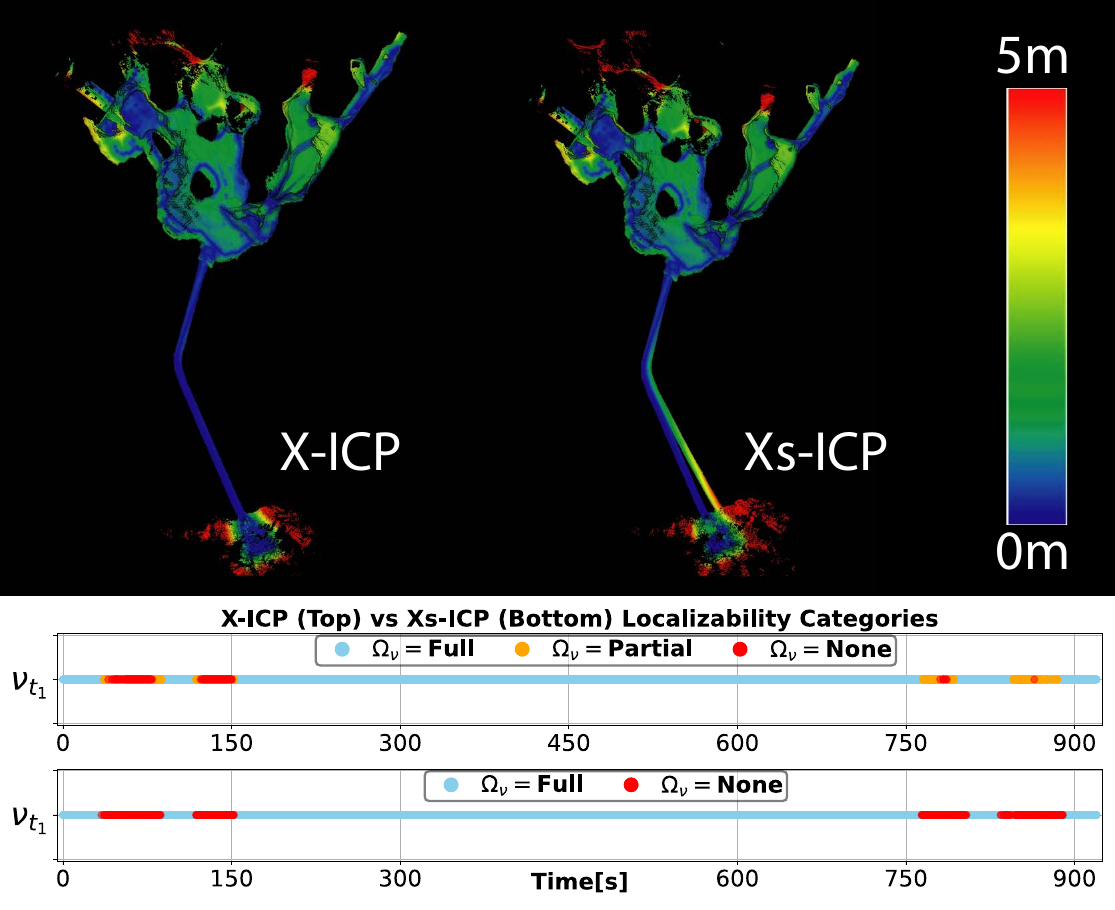}
\caption{\textbf{Top Row:} Point cloud maps of the proposed \mbox{X-ICP} and \mbox{Xs-ICP} for the Seem\"uhle experiment, using the \mbox{VLP-16} LiDAR. The color bar indicates the point-to-point distance error. \textbf{Bottom Row:} Predicted localizability categories of \mbox{X-ICP} and \mbox{Xs-ICP} for the tunnel section .}
\label{fig:seemhule_vlp16_ablation}
\vspace{-1.0em}
\end{figure}
\subsubsection{Seem\"uhle}
The two versions of the proposed method are compared on the Seem\"uhle experiment with \mbox{VLP-16} LiDAR data. The results are shown in Fig.~\ref{fig:seemhule_vlp16_ablation}. An increase in the robot pose drift is apparent when \textit{partial-localizability} is disabled in \mbox{Xs-ICP}, implying that using the three-level localizability detection can improve point cloud registration compared to two-level localizability detection. 
The localizability detection of the two variants is studied in the bottom plot of Fig.~\ref{fig:seemhule_vlp16_ablation}. In particular, when traversing the tunnel for the second time, the partial-localizability dominates the localizability category alongside the degenerate tunnel direction, while \mbox{Xs-ICP} estimates the localizability as non-localizable and fully relies on the prior for this eigenvector direction. 
Quantitative results of the RPE metric, shown in Table~\ref{table:RPE_results_xs_icp}, indicate that \mbox{Xs-ICP} performs comparably for rotation estimation while \mbox{X-ICP} performs significantly better for translation estimation. 
\begin{figure}[h]
\includegraphics[width=\linewidth]{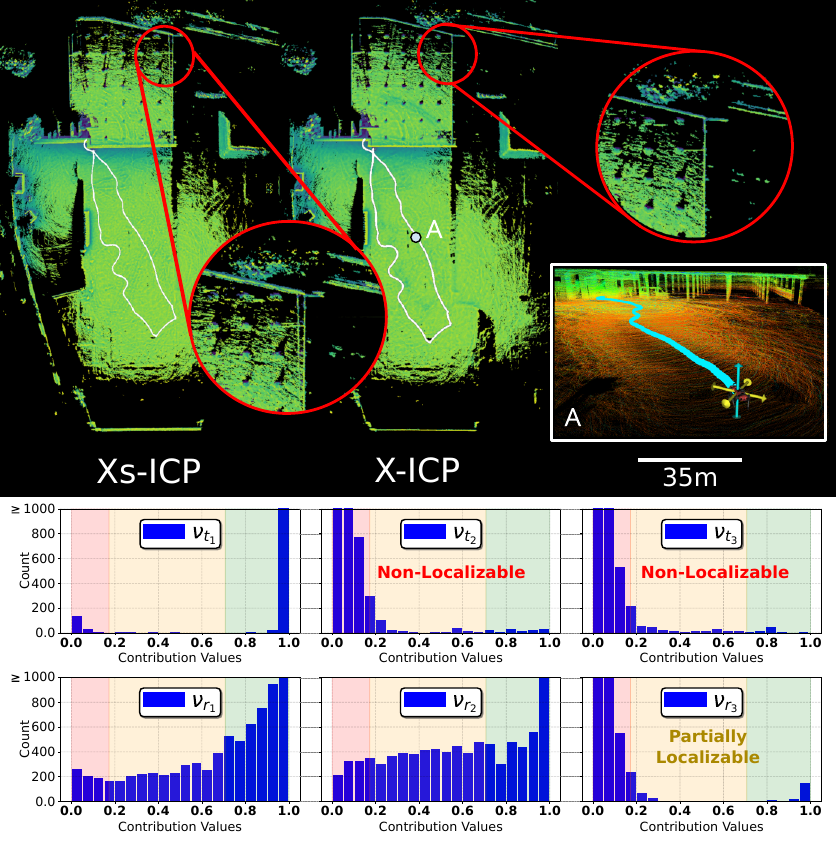}
\caption{\textbf{Top:} Point cloud maps of \mbox{X-ICP} and \mbox{Xs-ICP} approaches for the R\"umlang test site. A close-up of point \textbf{A} illustrates the detected degeneracy. \textbf{Bottom:} Contribution values are shown for point \textbf{A} as processed by \mbox{X-ICP}.}
\label{fig:rumlang_ablation}
\vspace{-1.0em}
\end{figure}
\begin{table}[h]
\caption{APE Error for \mbox{VLP-16} Seem\"uhle experiment  (best in \textbf{bold}).}
\resizebox{\columnwidth}{!}{
\begin{tabular}{lccccc}
\toprule[1pt]
\multirow{2}[3]{*}{} & \multicolumn{2}{c}{\makecell{First \SI{15}{\meter} Alignment}} & \multicolumn{3}{c}{Origin Alignment} \\
\cmidrule(lr){2-3}  \cmidrule(lr){4-6}
& \makecell{Translation \\ $\mu(\sigma)[m]$ } & \makecell{Rotation \\ $\mu(\sigma)[\deg]$} & \makecell{Translation \\ $\mu(\sigma)[m]$ } & \makecell{Rotation \\ $\mu(\sigma)[\deg]$} & \makecell{Last Position \\ Error[$m$]}\\
\toprule[1pt]
\makecell{X-ICP (Proposed)} &\textbf{2.05}(1.23) & \textbf{2.55}(\textbf{0.76}) & \textbf{2.45}(1.35) & \textbf{2.50}(\textbf{1.03}) & \textbf{0.27} \\[0.5em]
\makecell{Xs-ICP (Proposed)} &2.29(\textbf{1.22}) &3.06(1.19) & 2.68(\textbf{1.26}) & 3.08(1.3) & 5.34 \\[0.3em]
\toprule[1pt]
\end{tabular}
}
\label{table:APE_results_xs_icp}
\vspace{-1.0em}
\end{table}
\begin{table}[h!]\centering
\caption{RPE per \SI{10}{\meter} traversed distance for Seem\"uhle Mine experiment with \mbox{VLP-16} (best in \textbf{bold}).}
\begin{tabular}{lcc}
\toprule[1pt]
& \makecell{Translation \\ $\mu(\sigma)[m]$ } & \makecell{Rotation \\ $\mu(\sigma)[\deg]$}\\
\toprule[1pt]
\makecell{X-ICP (Proposed)} &\textbf{0.17}(\textbf{0.12}) &0.86(\textbf{0.42}) \\[0.5em]
\makecell{Xs-ICP (Proposed)} & 0.19(0.13) &\textbf{0.85}(0.47) \\[0.3em]
\toprule[1pt]
\end{tabular}
\label{table:RPE_results_xs_icp}
\vspace{-2.0em}
\end{table}
On the other hand, the APE metric and the end translation errors in Table~\ref{table:APE_results_xs_icp} indicate consistently better performance of \mbox{X-ICP}.

\subsubsection{R\"umlang}
A similar study is conducted for the R\"umlang experiment, refer to Fig.~\ref{fig:rumlang_ablation}. While the mapping performance of  \mbox{X-ICP} and  \mbox{Xs-ICP} is satisfactory, upon closer inspection, the point clouds registered for walls are duplicated for the \mbox{Xs-ICP} method, indicating a slight misalignment in rotation. 
Consequently, the partial category improves localizability estimation and mapping performance also for this case. A snapshot of the localizability analysis at point \textit{A} is shown in the lower part of Fig.~\ref{fig:rumlang_ablation}, illustrating the detected partial-localizability for one of the directions. 
The histograms show three directions; two of these directions are shown as yellow arrows in the bottom right image, which are the non-localizable translation directions, while the blue arrow indicates the partially-localizable rotation direction, justified by the few but highly informative information pairs seen at the bottom right histogram.

\begin{center}
\begin{figure}[h]
\includegraphics[trim=5 10 5 0, clip, width=\linewidth]{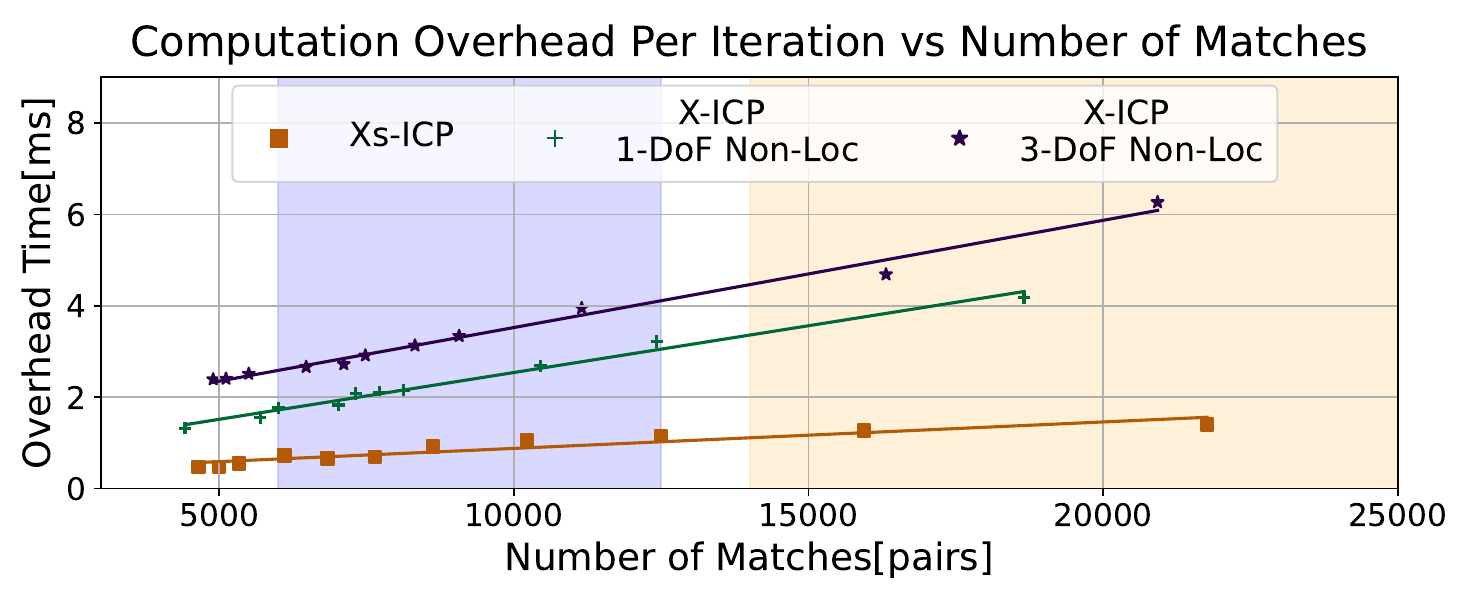}
\caption{\hspace{1ex}Scalability analysis of  \mbox{X-ICP} and \mbox{Xs-ICP}. The purple region (\protect\purpleline) and the light-orange region (\protect\greenlinet) span the number of possible matches for the \mbox{VLP-16} LiDAR and the Ouster \mbox{OS0-128} LiDAR, respectively.}
\label{fig:scalability_analysis}
\vspace{-1.0em}
\end{figure}
\end{center}

\subsubsection{Computational Time and Scalability}
Computational time and scalability analyses are performed to evaluate real-time applicability and the computational overhead of the proposed localizability aware registration methods. The scalability analysis in Fig.~\ref{fig:scalability_analysis} shows the overhead introduced by the localizability detection per ICP iteration, w.r.t. the number of correspondences of reading and reference point clouds. For \mbox{X-ICP}, the one- and three-axes degeneracy constitute the most common cases in real-world deployments (depicted in Fig~\ref{fig:scalability_analysis}). Since \mbox{Xs-ICP} does not require any re-sampling of points, the shown result represents all degeneracy conditions. As seen, both \mbox{X-ICP} and \mbox{Xs-ICP} methods are real-time capable for cases ranging from sparse 16-beam LiDARs to dense 128-beam LiDARs. By design, \mbox{Xs-ICP} requires less computation than \mbox{X-ICP}. Additionally, the real-time applicability of \mbox{X-ICP} and \mbox{Xs-ICP} is demonstrated by measuring the total scan-to-map registration time in the R\"umlang experiment on two computing platforms; one mobile and one desktop class system (Intel i7-9750H \& i9-13900K) are used. All evaluations shown in Table~\ref{table:computation} are performed in a single-threaded fashion for simplicity. The resulting computational analysis demonstrates the capability of \mbox{X-ICP} to run in real-time ($\mu=\SI{32.19}{\ms}$) on mobile robotic systems such as ANYmal. Moreover, in the absence of LiDAR degeneracy, the computational burden of \mbox{X-ICP} and \mbox{Xs-ICP} is negligible compared to the baseline.

\begin{table}[t]\centering
\caption{The scan-to-map registration time of different registration strategies for the R\"umlang dataset.}
\begin{threeparttable}
\begin{tabular}{lccc}
\toprule[1pt]
& \makecell{\mbox{X-ICP} \\ $\mu\:(\sigma)$ [ms]} & \makecell{\mbox{Xs-ICP} \\ $\mu\:(\sigma)$ [ms]} & \makecell{\mbox{${}^{*}_{}\text{Baseline}^{}_{}$} \\ $\mu\:(\sigma)$ [ms]}\\
\toprule[1pt]
\makecell{Intel i9-13900K} &{12.65} ({3.51}) & 11.14 ({3.1}) & {11.1} (1.45)\\[0.5em]
\makecell{Intel i7-9750H} &32.19 (10.7) & {29.42} (9.35) & {20.05} (2.21) \\
\toprule[1pt]
\end{tabular}
\begin{tablenotes}\footnotesize
\item[*] The baseline statistics are calculated until \SI{150}{\second} into the experiment to prevent the effect of degeneracy.
\end{tablenotes}
\end{threeparttable}
\label{table:computation}
\vspace{-1.0em}
\end{table}

\section{Conclusions and Future Work}\label{section:conclusion}
This work presented \mbox{X-ICP}, a localizability-aware LiDAR point cloud registration method, to enable robust and reliable pose estimation in challenging LiDAR degenerate environments. The proposed approach detects environmental degeneracy and calculates additional constraints for the ICP optimization problem. Adding these constraints to the optimization prohibits pose updates along degenerate directions. Moreover, by introducing partial localizability, the proposed method benefits from sparse but valuable information often present in real-world scenarios.
The efficacy of \mbox{X-ICP} is demonstrated through three real-world experiments and simulation-based analysis, all containing challenging environments causing ICP optimization to be ill-conditioned for accurate pose estimation. An ablation study further underlines the presented design choices. Currently, the proposed method is (similar to related methods) sensitive to the quality of the initial guess. In the future, this sensitivity of the point cloud registration method to initial guess quality should be reduced. Moreover, the sensor-dependent selection process of $\kappa_1$ is planned to be improved by introducing point-wise confidence filter values to the pairs as weights to reduce the effect of noise on the contribution calculation. Finally, the proposed fine-grained localizability will be integrated into a graph-based degeneracy-aware sensor fusion framework~\cite{nubertGraph} in the form of partial factors~\cite{hinduja2019degeneracy}.



\bibliographystyle{IEEEtran}
\bibliography{references.bib}

\begin{thebibliography}{10}
\providecommand{\url}[1]{#1}
\csname url@rmstyle\endcsname
\providecommand{\newblock}{\relax}
\providecommand{\bibinfo}[2]{#2}
\providecommand\BIBentrySTDinterwordspacing{\spaceskip=0pt\relax}
\providecommand\BIBentryALTinterwordstretchfactor{4}
\providecommand\BIBentryALTinterwordspacing{\spaceskip=\fontdimen2\font plus
\BIBentryALTinterwordstretchfactor\fontdimen3\font minus \fontdimen4\font\relax}
\providecommand\BIBforeignlanguage[2]{{%
\expandafter\ifx\csname l@#1\endcsname\relax
\typeout{** WARNING: IEEEtran.bst: No hyphenation pattern has been}%
\typeout{** loaded for the language `#1'. Using the pattern for}%
\typeout{** the default language instead.}%
\else
\language=\csname l@#1\endcsname
\fi
#2}}

\bibitem{slamOverview1}
R.~Latif, K.~Dahmane, and A.~Saddik, ``Slam algorithm: Overview and evaluation in a heterogeneous system,'' \emph{Enabling Machine Learning Applications in Data Science}, pp. 165--177, 2021.

\bibitem{cadena2016past}
C.~Cadena \emph{et~al.}, ``Past, present, and future of simultaneous localization and mapping: Toward the robust-perception age,'' \emph{IEEE Transactions on Robotics}, vol.~32, 2016.

\bibitem{slamOverview2}
K.~Ebadi \emph{et~al.}, ``Present and future of slam in extreme underground environments,'' \emph{arXiv preprint arXiv:2208.01787}, 2022.

\bibitem{geiger2012we}
A.~Geiger, P.~Lenz, and R.~Urtasun, ``Are we ready for autonomous driving? the kitti vision benchmark suite,'' in \emph{2012 IEEE conference on computer vision and pattern recognition}.\hskip 1em plus 0.5em minus 0.4em\relax IEEE, 2012, pp. 3354--3361.

\bibitem{helmberger2022hilti}
M.~Helmberger, K.~Morin, B.~Berner, N.~Kumar, G.~Cioffi, and D.~Scaramuzza, ``The hilti slam challenge dataset,'' \emph{IEEE Robotics and Automation Letters}, vol.~7, no.~3, pp. 7518--7525, 2022.

\bibitem{pointToPointICP}
P.~J. Besl and N.~D. McKay, ``Method for registration of 3-d shapes,'' in \emph{Sensor fusion IV: control paradigms and data structures}, vol. 1611.\hskip 1em plus 0.5em minus 0.4em\relax Spie, 1992, pp. 586--606.

\bibitem{point_to_plane}
K.-L. Low, ``Linear least-squares optimization for point-to-plane icp surface registration,'' \emph{Chapel Hill, University of North Carolina}, vol.~4, no.~10, pp. 1--3, 2004.

\bibitem{kiss_icp}
I.~Vizzo, T.~Guadagnino, B.~Mersch, L.~Wiesmann, J.~Behley, and C.~Stachniss, ``Kiss-icp: In defense of point-to-point icp--simple, accurate, and robust registration if done the right way,'' \emph{arXiv preprint arXiv:2209.15397}, 2022.

\bibitem{jelavic2022open3d}
E.~Jelavic, J.~Nubert, and M.~Hutter, ``Open3d slam: Point cloud based mapping and localization for education,'' in \emph{Robotic Perception and Mapping: Emerging Techniques, ICRA 2022 Workshop}.\hskip 1em plus 0.5em minus 0.4em\relax ETH Zurich, Robotic Systems Lab, 2022, p.~24.

\bibitem{censi}
A.~Censi, ``An accurate closed-form estimate of icp's covariance,'' in \emph{Proceedings 2007 IEEE international conference on robotics and automation}.\hskip 1em plus 0.5em minus 0.4em\relax IEEE, 2007, pp. 3167--3172.

\bibitem{brossard}
M.~Brossard, S.~Bonnabel, and A.~Barrau, ``A new approach to 3d icp covariance estimation,'' \emph{IEEE Robotics and Automation Letters}, vol.~5, no.~2, pp. 744--751, 2020.

\bibitem{solRemap}
J.~Zhang, M.~Kaess, and S.~Singh, ``On degeneracy of optimization-based state estimation problems,'' in \emph{2016 IEEE International Conference on Robotics and Automation (ICRA)}.\hskip 1em plus 0.5em minus 0.4em\relax IEEE, 2016, pp. 809--816.

\bibitem{selfSupervisedOdom}
J.~Nubert, S.~Khattak, and M.~Hutter, ``Self-supervised learning of lidar odometry for robotic applications,'' in \emph{2021 IEEE International Conference on Robotics and Automation (ICRA)}, 2021, pp. 9601--9607.

\bibitem{nubert2022learning}
J.~Nubert, E.~Walther, S.~Khattak, and M.~Hutter, ``Learning-based localizability estimation for robust lidar localization,'' in \emph{IEEE International Conference on Intelligent Robots and Systems (IROS)}, 2022.

\bibitem{zhen2017robust}
W.~Zhen, S.~Zeng, and S.~Soberer, ``Robust localization and localizability estimation with a rotating laser scanner,'' in \emph{IEEE International Conference on Robotics and Automation (ICRA)}, 2017.

\bibitem{tunnelLocalizability}
W.~Zhen and S.~Scherer, ``Estimating the localizability in tunnel-like environments using lidar and uwb,'' in \emph{2019 International Conference on Robotics and Automation (ICRA)}.\hskip 1em plus 0.5em minus 0.4em\relax IEEE, 2019, pp. 4903--4908.

\bibitem{hinduja2019degeneracy}
A.~Hinduja, B.-J. Ho, and M.~Kaess, ``Degeneracy-aware factors with applications to underwater slam,'' in \emph{IEEE International Conference on Intelligent Robots and Systems (IROS)}, 2019.

\bibitem{NDT}
P.~Biber and W.~Stra{\ss}er, ``The normal distributions transform: A new approach to laser scan matching,'' in \emph{IEEE International Conference on Intelligent Robots and Systems (IROS)}, 2003.

\bibitem{generalizedICP}
A.~Segal, D.~Haehnel, and S.~Thrun, ``Generalized-icp.'' in \emph{Robotics: science and systems}, vol.~2, no.~4.\hskip 1em plus 0.5em minus 0.4em\relax Seattle, WA, 2009, p. 435.

\bibitem{loam}
J.~Zhang and S.~Singh, ``Loam: Lidar odometry and mapping in real-time.'' in \emph{Robotics: Science and Systems}, vol.~2, 2014.

\bibitem{behley2018efficient}
J.~Behley and C.~Stachniss, ``Efficient surfel-based slam using 3d laser range data in urban environments.'' in \emph{Robotics: Science and Systems}, vol. 2018, 2018, p.~59.

\bibitem{pomerleau2013applied}
F.~Pomerleau, ``Applied registration for robotics: Methodology and tools for icp-like algorithms,'' Ph.D. dissertation, ETH Zurich, 2013.

\bibitem{pointToLine}
A.~Censi, ``An icp variant using a point-to-line metric,'' in \emph{IEEE International Conference on Robotics and Automation}, 2008.

\bibitem{point_to_gaussian}
P.~Babin, P.~Dandurand, V.~Kubelka, P.~Gigu{\`e}re, and F.~Pomerleau, ``Large-scale 3d mapping of subarctic forests,'' in \emph{Field and Service Robotics}.\hskip 1em plus 0.5em minus 0.4em\relax Springer, 2021, pp. 261--275.

\bibitem{symmetricPointToPlane}
S.~Rusinkiewicz, ``A symmetric objective function for icp,'' \emph{ACM Transactions on Graphics (TOG)}, vol.~38, no.~4, pp. 1--7, 2019.

\bibitem{litamin2}
M.~Yokozuka, K.~Koide, S.~Oishi, and A.~Banno, ``Litamin2: Ultra light lidar-based slam using geometric approximation applied with kl-divergence,'' in \emph{IEEE International Conference on Robotics and Automation (ICRA)}, 2021.

\bibitem{wildCat}
M.~Ramezani, K.~Khosoussi, G.~Catt, P.~Moghadam, J.~Williams, P.~Borges, F.~Pauling, and N.~Kottege, ``Wildcat: Online continuous-time 3d lidar-inertial slam,'' \emph{arXiv preprint arXiv:2205.12595}, 2022.

\bibitem{compSLAM}
S.~Khattak \emph{et~al.}, ``Complementary multi--modal sensor fusion for resilient robot pose estimation in subterranean environments,'' in \emph{IEEE International Conference on Unmanned Aircraft Systems (ICUAS)}, 2020.

\bibitem{dareSLAM}
K.~Ebadi, M.~Palieri, S.~Wood, C.~Padgett, and A.-a. Agha-mohammadi, ``Dare-slam: Degeneracy-aware and resilient loop closing in perceptually-degraded environments,'' \emph{Journal of Intelligent \& Robotic Systems}, vol. 102, no.~1, pp. 1--25, 2021.

\bibitem{teaser}
H.~Yang, J.~Shi, and L.~Carlone, ``Teaser: Fast and certifiable point cloud registration,'' \emph{IEEE Transactions on Robotics}, vol.~37, 2020.

\bibitem{wahba}
H.~Yang and L.~Carlone, ``A quaternion-based certifiably optimal solution to the wahba problem with outliers,'' in \emph{Proceedings of the IEEE/CVF International Conference on Computer Vision}, 2019, pp. 1665--1674.

\bibitem{lagrangianDuality}
L.~Carlone, D.~M. Rosen, G.~Calafiore, J.~J. Leonard, and F.~Dellaert, ``Lagrangian duality in 3d slam: Verification techniques and optimal solutions,'' in \emph{IEEE International Conference on Intelligent Robots and Systems (IROS)}, 2015.

\bibitem{chebrolu}
N.~Chebrolu, T.~L{\"a}be, O.~Vysotska, J.~Behley, and C.~Stachniss, ``Adaptive robust kernels for non-linear least squares problems,'' \emph{IEEE Robotics and Automation Letters}, vol.~6, no.~2, pp. 2240--2247, 2021.

\bibitem{graduated}
H.~Yang, P.~Antonante, V.~Tzoumas, and L.~Carlone, ``Graduated non-convexity for robust spatial perception: From non-minimal solvers to global outlier rejection,'' \emph{IEEE Robotics and Automation Letters}, 2020.

\bibitem{CELLO3D}
D.~Landry, F.~Pomerleau, and P.~Giguere, ``Cello-3d: Estimating the covariance of icp in the real world,'' in \emph{2019 International Conference on Robotics and Automation (ICRA)}.\hskip 1em plus 0.5em minus 0.4em\relax IEEE, 2019, pp. 8190--8196.

\bibitem{deepICP}
A.~De~Maio and S.~Lacroix, ``Deep bayesian icp covariance estimation,'' \emph{arXiv preprint arXiv:2202.11607}, 2022.

\bibitem{bonnabel2016covariance}
S.~Bonnabel, M.~Barczyk, and F.~Goulette, ``On the covariance of icp-based scan-matching techniques,'' in \emph{2016 American Control Conference (ACC)}.\hskip 1em plus 0.5em minus 0.4em\relax IEEE, 2016, pp. 5498--5503.

\bibitem{gelfand2003}
N.~Gelfand, L.~Ikemoto, S.~Rusinkiewicz, and M.~Levoy, ``Geometrically stable sampling for the icp algorithm,'' in \emph{IEEE International Conference on 3-D Digital Imaging and Modeling}, 2003.

\bibitem{kwok2016improvements}
T.-H. Kwok and K.~Tang, ``Improvements to the iterative closest point algorithm for shape registration in manufacturing,'' \emph{Journal of Manufacturing Science and Engineering}, vol. 138, no.~1, 2016.

\bibitem{deschaud2018imls}
J.-E. Deschaud, ``Imls-slam: Scan-to-model matching based on 3d data,'' in \emph{IEEE International Conference on Robotics and Automation}, 2018.

\bibitem{gramian}
Z.~Rong and N.~Michael, ``Detection and prediction of near-term state estimation degradation via online nonlinear observability analysis,'' in \emph{IEEE International Symposium on Safety, Security, and Rescue Robotics (SSRR)}, 2016.

\bibitem{planeRank}
H.~Cho, S.~Yeon, H.~Choi, and N.~Doh, ``Detection and compensation of degeneracy cases for imu-kinect integrated continuous slam with plane features,'' \emph{Sensors}, vol.~18, no.~4, p. 935, 2018.

\bibitem{LION}
A.~Tagliabue \emph{et~al.}, ``Lion: Lidar-inertial observability-aware navigator for vision-denied environments,'' in \emph{International Symposium on Experimental Robotics}, 2020.

\bibitem{fisher1}
Y.~Liu, J.~Wang, and Y.~Huang, ``A localizability estimation method for mobile robots based on 3d point cloud feature,'' in \emph{IEEE International Conference on Real-time Computing and Robotics (RCAR)}, 2021.

\bibitem{fisher2}
L.~Dong, W.~Chen, and J.~Wang, ``Efficient feature extraction and localizability based matching for lidar slam,'' in \emph{IEEE International Conference on Robotics and Biomimetics (ROBIO)}, 2021.

\bibitem{rankDeficientSLAM}
S.~B. Nashed, J.~J. Park, R.~Webster, and J.~W. Durham, ``Robust rank deficient slam,'' in \emph{IEEE/RSJ International Conference on Intelligent Robots and Systems (IROS)}.\hskip 1em plus 0.5em minus 0.4em\relax IEEE, 2021, pp. 6603--6608.

\bibitem{dOpt}
M.~L.~R. Ar{\'e}valo, ``On the uncertainty in active slam: representation, propagation and monotonicity,'' Ph.D. dissertation, Universidad de Zaragoza, 2018.

\bibitem{M_LOAM}
J.~Jiao, H.~Ye, Y.~Zhu, and M.~Liu, ``Robust odometry and mapping for multi-lidar systems with online extrinsic calibration,'' \emph{IEEE Transactions on Robotics}, 2021.

\bibitem{solRemap_kaess}
E.~Westman and M.~Kaess, ``Degeneracy-aware imaging sonar simultaneous localization and mapping,'' \emph{IEEE Journal of Oceanic Engineering}, vol.~45, no.~4, pp. 1280--1294, 2019.

\bibitem{ren2021towards}
R.~Ren, H.~Fu, H.~Xue, Z.~Sun, K.~Ding, and P.~Wang, ``Towards a fully automated 3d reconstruction system based on lidar and gnss in challenging scenarios,'' \emph{Remote Sensing}, vol.~13, no.~10, p. 1981, 2021.

\bibitem{zhou2020lidar}
H.~Zhou, Z.~Yao, and M.~Lu, ``Lidar/uwb fusion based slam with anti-degeneration capability,'' \emph{IEEE Transactions on Vehicular Technology}, vol.~70, no.~1, pp. 820--830, 2020.

\bibitem{nobili1}
S.~Nobili, G.~Tinchev, and M.~Fallon, ``Predicting alignment risk to prevent localization failure,'' in \emph{2018 IEEE International Conference on Robotics and Automation (ICRA)}.\hskip 1em plus 0.5em minus 0.4em\relax IEEE, 2018, pp. 1003--1010.

\bibitem{overlapnet}
X.~Chen \emph{et~al.}, ``Overlapnet: A siamese network for computing lidar scan similarity with applications to loop closing and localization,'' \emph{Autonomous Robots}, 2022.

\bibitem{deepLocalizability}
Y.~Gao, S.~Q. Wang, J.~H. Li, M.~Q. Hu, H.~Y. Xia, H.~Hu, and L.~J. Wang, ``A prediction method of localizability based on deep learning,'' \emph{IEEE Access}, vol.~8, pp. 110\,103--110\,115, 2020.

\bibitem{flory2009constrained}
S.~Floery, ``Constrained matching of point clouds and surfaces (ph. d. thesis),'' \emph{Technische Universitt Wien}, 2010.

\bibitem{constraintOptimization}
C.~Olsson and A.~Eriksson, ``Solving quadratically constrained geometrical problems using lagrangian duality,'' in \emph{2008 19th International Conference on Pattern Recognition}.\hskip 1em plus 0.5em minus 0.4em\relax IEEE, 2008, pp. 1--5.

\bibitem{sparseICP}
S.~Bouaziz, A.~Tagliasacchi, and M.~Pauly, ``Sparse iterative closest point,'' in \emph{Computer graphics forum}, vol.~32, no.~5.\hskip 1em plus 0.5em minus 0.4em\relax Wiley Online Library, 2013, pp. 113--123.

\bibitem{ct_icp}
P.~Dellenbach, J.-E. Deschaud, B.~Jacquet, and F.~Goulette, ``Ct-icp: Real-time elastic lidar odometry with loop closure,'' in \emph{IEEE International Conference on Robotics and Automation (ICRA)}, 2022.

\bibitem{svd}
M.~E. Wall, A.~Rechtsteiner, and L.~M. Rocha, ``Singular value decomposition and principal component analysis,'' in \emph{A practical approach to microarray data analysis}.\hskip 1em plus 0.5em minus 0.4em\relax Springer, 2003, pp. 91--109.

\bibitem{ICPderivation}
F.~Pomerleau, F.~Colas, R.~Siegwart, \emph{et~al.}, ``A review of point cloud registration algorithms for mobile robotics,'' \emph{Foundations and Trends{\textregistered} in Robotics}, vol.~4, no.~1, pp. 1--104, 2015.

\bibitem{pharosHiltiReport}
\BIBentryALTinterwordspacing
Y.~Nava \emph{et~al.}, ``Lidar-inertial odometry,'' ANYbotics,'' Report, 2022. [Online]. Available: \url{http://hdl.handle.net/20.500.11850/580200}
\BIBentrySTDinterwordspacing

\bibitem{libpointmatcher}
F.~Pomerleau, F.~Colas, R.~Siegwart, and S.~Magnenat, ``Comparing icp variants on real-world data sets,'' \emph{Autonomous Robots}, vol.~34, 2013.

\bibitem{DNSS}
T.-H. Kwok, ``Dnss: Dual-normal-space sampling for 3-d icp registration,'' \emph{IEEE Transactions on Automation Science and Engineering}, vol.~16, no.~1, pp. 241--252, 2018.

\bibitem{lu}
C.~Fu, X.~Jiao, and T.~Yang, ``Efficient sparse lu factorization with partial pivoting on distributed memory architectures,'' \emph{IEEE Transactions on Parallel and Distributed Systems}, vol.~9, no.~2, pp. 109--125, 1998.

\bibitem{RIF}
M.~Benzi and M.~T{\u{u}}ma, ``A robust incomplete factorization preconditioner for positive definite matrices,'' \emph{Numerical Linear Algebra with Applications}, vol.~10, no. 5-6, pp. 385--400, 2003.

\bibitem{lagrangianMultipliers}
D.~P. Bertsekas, \emph{Constrained optimization and Lagrange multiplier methods}.\hskip 1em plus 0.5em minus 0.4em\relax Academic press, 2014.

\bibitem{RTC360}
A.~Biasion, T.~Moerwald, B.~Walser, and G.~Walsh, ``A new approach to the terrestrial laser scanner workflow: the rtc360 solution,'' \emph{Geospatial Information for a Smarter Life and Environmental Resilience}, 2019.

\bibitem{ANYmal}
M.~Hutter \emph{et~al.}, ``Anymal-toward legged robots for harsh environments,'' \emph{Advanced Robotics}, vol.~31, 2017.

\bibitem{tsif}
M.~Bloesch, M.~Burri, H.~Sommer, R.~Siegwart, and M.~Hutter, ``The two-state implicit filter recursive estimation for mobile robots,'' \emph{IEEE Robotics and Automation Letters}, vol.~3, no.~1, pp. 573--580, 2017.

\bibitem{newerCollegeDataset}
M.~Ramezani \emph{et~al.}, ``The newer college dataset: Handheld lidar, inertial and vision with ground truth,'' in \emph{IEEE International Conference on Intelligent Robots and Systems (IROS)}, 2020.

\bibitem{ate_rpe}
J.~Sturm, N.~Engelhard, F.~Endres, W.~Burgard, and D.~Cremers, ``A benchmark for the evaluation of rgb-d slam systems,'' in \emph{IEEE international conference on intelligent robots and systems}, 2012.

\bibitem{nubertGraph}
J.~Nubert, S.~Khattak, and M.~Hutter, ``Graph-based multi-sensor fusion for consistent localization of autonomous construction robots,'' in \emph{International Conference on Robotics and Automation (ICRA)}, 2022.

\end{thebibliography}

\vspace{-1.5cm}

\begin{IEEEbiography}[{\includegraphics[trim={18 0 18 0},clip,width=1in,height=1.25in,keepaspectratio]{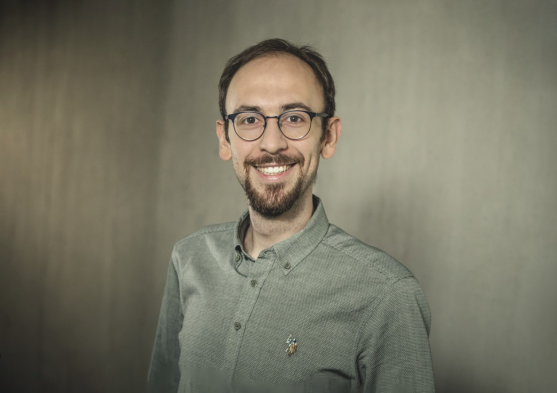}}]%
{Turcan Tuna} is a PhD student in the Robotic Systems Lab at ETH Zurich. He received his M.Sc. in Robotics, Systems \& Control in 2022 from ETH Zurich. Previously, he completed a double major, B.Sc in Mechanical Engineering and Control \& Automation Engineering, at Istanbul Technical University. He graduated from both of his B.Sc majors with distinction. His research interests include developing and deploying robust localization, perception, and mapping frameworks on robotic systems.
\end{IEEEbiography}

\vspace{-1.5cm}

\begin{IEEEbiography}[{\includegraphics[trim={236 0 236 0},clip,width=1in,height=1.25in,keepaspectratio]{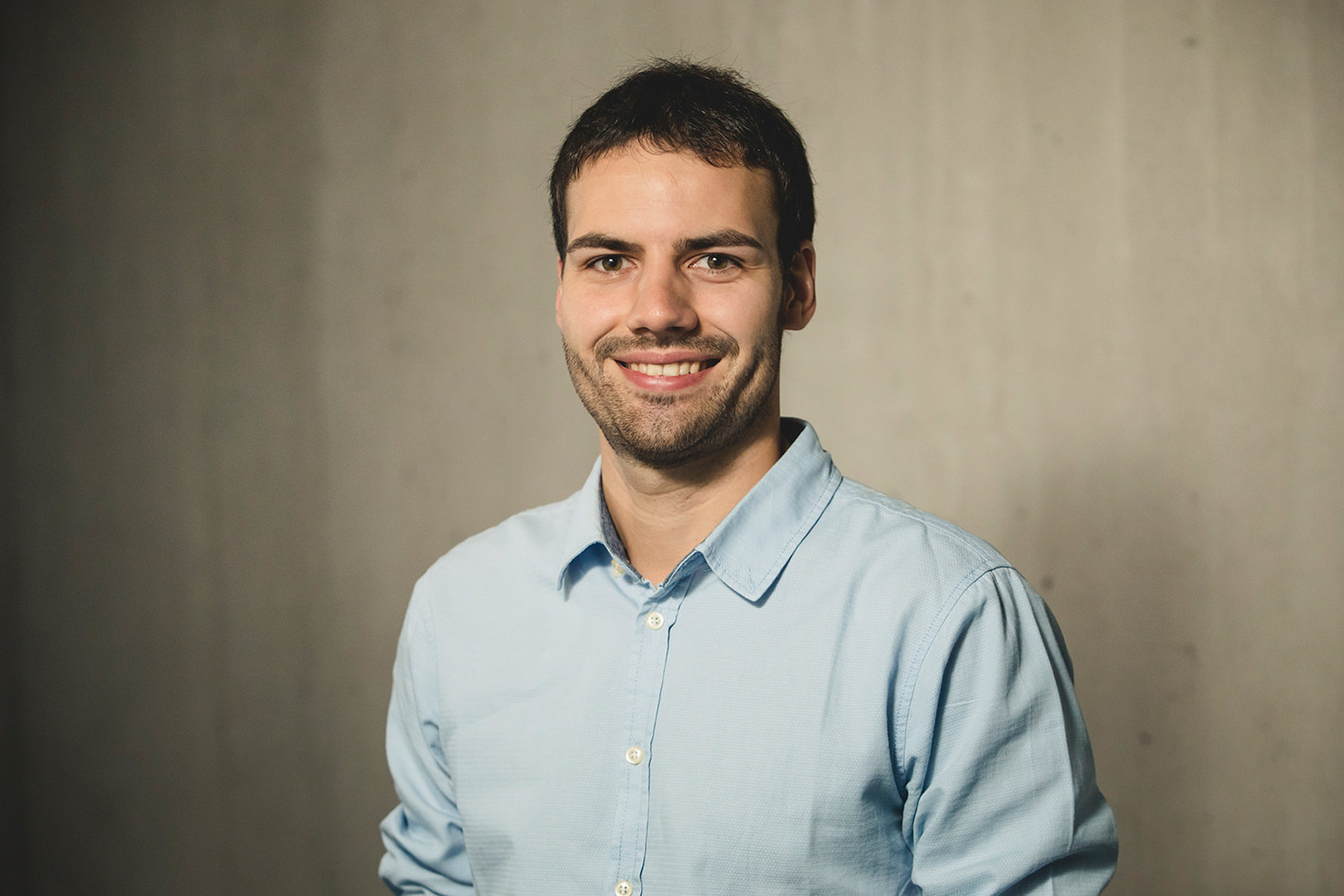}}]%
{Julian Nubert} is a PhD student in the Robotic Systems Lab at ETH Zurich. He received his M.Sc. in Robotics, Systems \& Control in 2020 from ETH Zurich. He is also affiliated with the Max Planck Institute through the MPI ETH Center for Learning Systems. His research interests lie in the field of robust robot perception, and how it can be used for the deployment of mobile robotic systems. Julian received the ETH silver medal and was awarded the Willi-Studer-Price for his accomplishments during his master studies.
\end{IEEEbiography}

\vspace{-1.5cm}

\begin{IEEEbiography}[{\includegraphics[trim={18 0 18 0},clip,width=1in,height=1.25in,keepaspectratio]{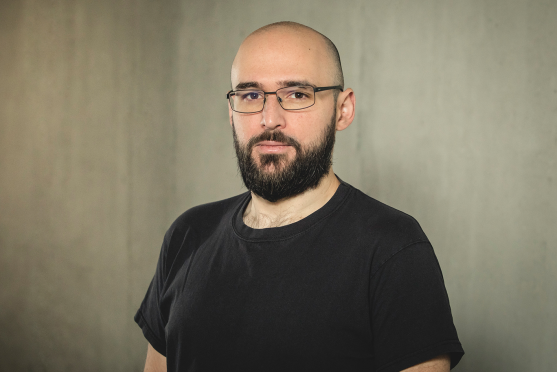}}]%
{Yoshua Nava} is a Perception Software Engineer at ANYbotics AG (Zurich, Switzerland), where he specializes in point cloud-based localization and mapping. He received his masters degree in Systems, Control and Robotics from KTH Royal Institute of Technology (Stockholm, Sweden), and his bachelor's degree from Universidad Católica Andrés Bello (Caracas, Venezuela). His research interests broadly cover localization and mapping as a core skill for robot mobility, and as a way to increase situational awareness for robots and human operators.
\end{IEEEbiography}

\vspace{-1.5cm}

\begin{IEEEbiography}[{\includegraphics[trim={236 0 236 0}, clip,width=1in,height=1.25in,keepaspectratio]{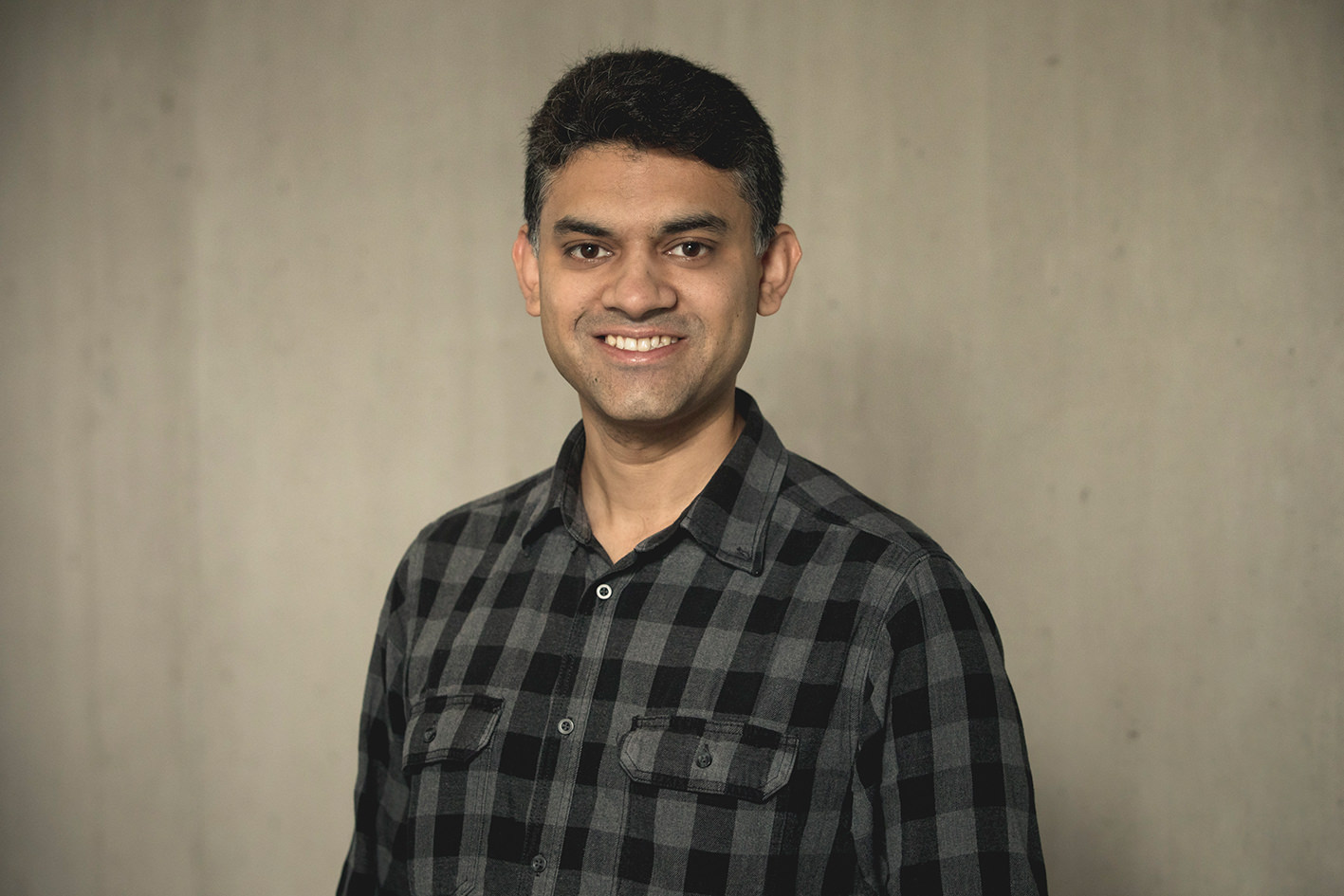}}]%
{Shehryar Khattak} is a currently a Robotics Technologist at the NASA - Jet Propulsion Lab. Previously, he was a post-doctoral researcher at the Robotics Systems Lab at ETH Zurich and the perception lead of team CERBERUS, which won the DARPA SubT Challenge (2021). He received his Ph.D. (2019) and MS (2017) in Computer Science from the University of Nevada, Reno. He also holds an MS in Aerospace Engineering from KAIST (2012) and a BS in Mechanical Engineering from GIKI (2009). His research focuses on developing perception algorithms to support real-time localization and mapping for field robotics applications.
\end{IEEEbiography}

\vspace{-1.5cm}

\begin{IEEEbiography}[{\includegraphics[trim={236 0 236 0}, clip,width=1in,height=1.25in,keepaspectratio]{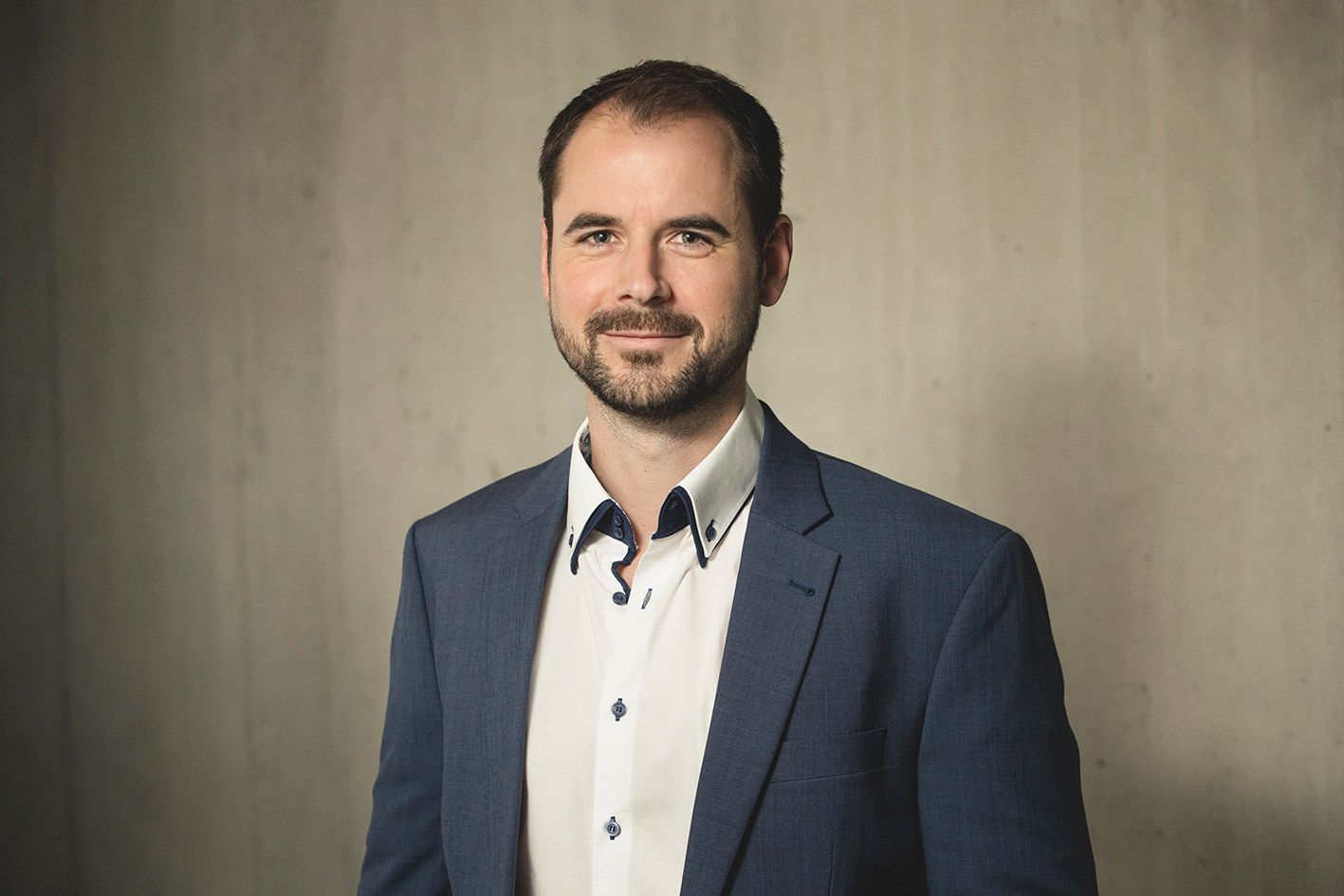}}]%
{Marco Hutter} is Associate Professor for Robotic Systems at ETH Zurich. He received his M.Sc. and PhD from ETH Zurich in 2009 and 2013 in the field of design, actuation, and control of legged robots. His research interests are in the development of novel machines and actuation concepts together with the underlying control, planning, and machine learning algorithms for locomotion and manipulation. Marco is the recipient of an ERC Starting Grant, PI of the NCCRs robotics and digital fabrication, PI in various EU projects and international challenges, a co-founder of several ETH Startups such as ANYbotics AG.
\end{IEEEbiography}

\end{document}